\documentclass[11pt]{article}

\usepackage[preprint]{acl}

\usepackage{times}
\usepackage{latexsym}
\usepackage{hyperref}
\usepackage{multirow}

\usepackage[T1]{fontenc}

\usepackage[utf8]{inputenc}

\usepackage{microtype}

\usepackage{inconsolata}

\usepackage{graphicx}

\usepackage{subcaption}

\usepackage{booktabs}

\newcommand{\derelab}{\textsc{DeReLab}}

\usepackage{algorithm}
\usepackage{algpseudocode}
\usepackage{amsmath}
\usepackage{tcolorbox}

\newtcolorbox{promptbox}[1]{
    colback= gray !5,
    colframe=orange !70,
    fonttitle=\bfseries,
    title=#1,
    arc=2mm,
    left=10pt,
    right=10pt,
    top=10pt,
    bottom=10pt
}
\title{\derelab: Probing Defeasible Reasoning and Confirmation Bias in LLMs with a Generative Benchmark}
\author{
Jayanta Sadhu$^{1}$,
Sayem Shahad$^{2}$,
Kenneth Marino$^{1}$ \\
$^{1}$University of Utah \\
$^{2}$Bangladesh University of Engineering and Technology \\
\texttt{\{jayanta.sadhu, kenneth.marino\}@utah.edu} \\
\texttt{1024052110@grad.cse.buet.ac.bd}
}

\begin{document}
\maketitle
\begin{abstract}
Defeasible reasoning is a type of reasoning where inferences are drawn from plausible current evidence, but can be retracted upon the introduction of newer evidence. Although recent studies have examined language-model behaviors in defeasible reasoning, the datasets have been static and lack wide coverage of non-monotonic reasoning categories. We introduce \derelab{}, a generative framework 
that produces multi-turn belief-updating 
conversations from parameterized graph structures across 
default and inheritance reasoning, with formally verified 
ground truth at every turn, enabling controlled measurement of how 
models respond to confirming and disconfirming evidence. This controlled generation process creates a testbed for experimental 
designs that isolate specific reasoning demands.  Applying this capability to the study of confirmation bias, we evaluate nine open and proprietary large language models and find that nearly all exhibit a systematic 
tendency to accept congruent evidence while resisting 
incongruent updates, with several models correctly 
identifying a weakening update yet failing to revise their 
conclusion. We believe our work and findings will facilitate future research on evaluating language models in defeasible reasoning. \footnote{All our data and code are public and available at \href{https://github.com/Jayanta47/DeReLab}{\texttt{https://github.com/Jayanta47/DeReLab}}} 
\end{abstract}

\section{Introduction}

Reasoning is a central capability for any intelligent agent, and it plays a decisive role in performing reliably in real-world applications. The majority of inferences humans make in real life are defeasible in nature \citep{CHATER2011553}. Unlike controlled environments, practical reasoning is rarely straightforward due to noisy data and incomplete information flow. With the integration of LLMs into many of our day-to-day activities, it has become important to study language model behaviors for defeasible reasoning, which has been relatively understudied to date. 

Prior works on defeasible inference have evaluated language
models using static datasets through multiple probing methods.
\citet{rudinger-etal-2020-thinking} created the first
defeasible reasoning natural language dataset and
\citet{allaway-mckeown-2025-evaluating} created a dataset
grounded in generics and inheritance properties.
These datasets are, however, susceptible to benchmark saturation as model capabilities improve. With the rapid growth of the field, evaluation
datasets often become short-lived, and researchers must
handcraft new versions or apply targeted modifications.
Furthermore, static benchmarks report aggregate accuracy
without revealing \emph{why} a model fails or whether
failures reflect genuine reasoning deficits rather than
surface-level pattern matching.

We address these limitations with \derelab{} (\textbf{De}feasible
\textbf{Re}asoning \textbf{Lab}), a generative framework that
produces multi-turn belief-updating conversations from
parameterized graph structures across two canonical
non-monotonic reasoning paradigms --- default reasoning and
inheritance reasoning~\citep{10.5555/73682.73696, HORTY1990311}
--- with formally verified ground truth at every turn.
\derelab{} follows the generative evaluation paradigm
of~\citet{ICLR2025_6fc46679} and extends it to defeasible
reasoning, a fundamentally different setting that requires
models to retract conclusions when new evidence defeats
prior defaults --- a capability that relational reasoning
benchmarks do not test.
Fine-grained configuration parameters over property chaining,
branching factor, distractor density, and source-priority structure enable controlled diagnostic experiments at
arbitrary difficulty levels without collecting a new dataset.

\begin{figure*}[t]
    \centering
    \includegraphics[width = 1.0\linewidth]{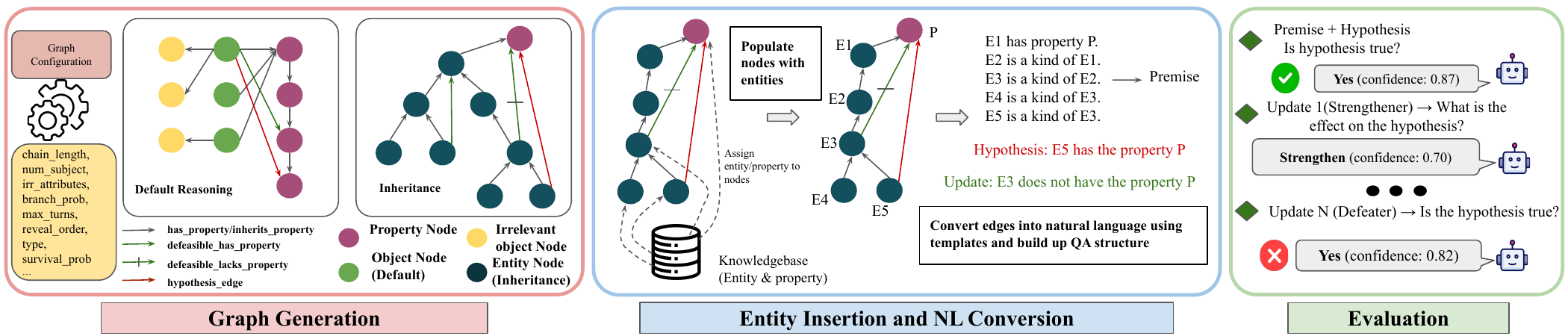}
    \caption{Overview of the generation and evaluation process of \derelab{}.}
    \label{fig:derelab_structure}
\end{figure*}

Using \derelab{}, we evaluate nine base models under ten model-inference configurations and report three findings.
\emph{First}, no model achieves strong performance across
all conditions: defeasible reasoning is genuinely hard,
and accuracy degrades systematically as reasoning
complexity increases.
\emph{Second}, confirmation bias~\citep{nickerson1998confirmation}
--- the tendency to accept evidence that confirms a prior
belief while resisting evidence that contradicts it ---
is near-universal among models on default reasoning,
revealing that LLMs exhibit a well-documented human
cognitive bias when updating beliefs under new evidence.
\emph{Third}, several models correctly identify that an
update weakens the hypothesis yet still fail to revise
their conclusion --- suggesting that belief revision and not factual understanding is the locus of failure.

In this paper, we: (1) introduce \derelab{}, a flexible dataset generation framework for scalable evaluation and cognitive science-inspired probes of LLMs on defeasible reasoning;
(2) demonstrate systematic evaluation across default and
inheritance reasoning at multiple difficulty tiers, revealing
that reasoning complexity and task type jointly determine
model failure; and (3) show how the framework enables
cognitive science-inspired probes, uncovering near-universal
confirmation bias and a know-but-don't-output dissociation
across current LLMs.

\section{Related Work}
\paragraph{Defeasible and Non-monotonic Reasoning}
Preliminary studies on AI and defeasible reasoning focused on formal logics (e.g., \citet{Reiter1978, REITER198081, POOLE198827}). In recent times, research has been done to construct datasets built upon these principles and evaluate non-monotonic reasoning ability in language models. \citet{xiu-etal-2022-logicnmr} constructed a dataset for non-monotonic proofs and \citet{parmar-etal-2024-logicbench} used patterns of default reasoning from \citet{10.5555/73682.73696} to construct a QA dataset. Non-monotonic reasoning has also been studied within the context of natural language inference (NLI) \citep{yanaka-etal-2019-help, yanaka-etal-2019-neural, 10.1007/s10849-023-09410-4}. Although these datasets focused on evaluating non-monotonic reasoning, these formulations did not support incremental information updates. Subsequent work (e.g., \citet{zhao-etal-2023-abductive, rudinger-etal-2020-thinking}) addressed this limitation by introducing additional information that can alter an inference. Evaluating the defeasibility of reasoning is a more nuanced way of evaluating non-monotonic reasoning. A formal modal framework for defeasibility in non-monotonic reasoning was proposed by \citet{10.1007/BFb0018430}. \citet{allaway-etal-2023-penguins} discussed generics and instantiations in logical programming, and \citet{allaway-mckeown-2025-evaluating} created a dataset for defeasible reasoning with the inheritance reasoning category. 

\paragraph{Cognitive biases in LLMs} 
The use of Large Language Models (LLMs) in high-stakes decision-making (e.g., \citet{wu2023bloomberggptlargelanguagemodel, singhal2022largelanguagemodelsencode}) has made it imperative to broaden the study of biases beyond traditional ethical and social concerns (e.g., \citet{Gallegos2023BiasAF}). Specifically, there is a growing need to account for cognitive biases and heuristics, which directly impact the rationality of LLMs' judgments (\citet{Hagendorff2022HumanlikeIB}). Early research in this area focused on detecting these effects at the level of individual prompts (\citet{talboy2023challengingappearancemachineintelligence, macmillanscott2024irrationalitycognitivebiaseslarge}). Other works have investigated the challenge of detection and mitigation, but were limited to particular LLM roles (\citet{pilli2024exploringconversationalagentseffective, ye2025justice}), or specific domains (\citet{schmidgall2024addressingcognitivebiasmedical, opedal2024cognitivebias}). To address the need for large-scale evaluation, follow-up works have proposed comprehensive frameworks, such as those by \citet{echterhoff-etal-2024-cognitive} and \citet{Xie2024MindScopeEC}, which aim to systematically benchmark and mitigate these cognitive limitations (e.g., \citet{Zhong2024BalancingRA}).

\paragraph{Sandbox Evaluation Toolkit}
To take evaluation beyond simple pattern matching, \citet{clark2020transformers} introduced the concept of transformers emulating a reasoning algorithm by concluding explicit rules provided in natural language. \citet{tafjord2021proofwriter} extended it by enabling iterative 1-hop generation to build a proof reflecting the model's decisions. Later, \citet{kiela-etal-2021-dynabench} and \citet{thrush2022dynatask} added human annotation in the evaluation loop to address the rapid saturation of static benchmarks. Recognizing human annotation as a bottleneck, \citet{ICLR2025_6fc46679} proposed a framework to generate reasoning datasets given tunable input configurations and seeds. 
Our framework draws inspiration from their work, but significantly differs in the generation and evaluation paradigms.


\section{The \textsc{DeReLab} Framework}
\label{sec:framework}

\textsc{DeReLab} is a generative evaluation framework for defeasible reasoning that overcomes three limitations of existing benchmarks. \textit{First}, unlike static datasets that become obsolete as model capabilities advance~\citep{kiela-etal-2021-dynabench}, \textsc{DeReLab} generates evaluation problems on demand from a scalable configuration space, allowing difficulty to grow with model capability without 
collecting new data.
\textit{Second}, unlike prior defeasible datasets that vary primarily at the semantic level, \textsc{DeReLab} spans a rich distribution of reasoning \emph{structures} --- linear default chains, branching inheritance hierarchies, multi-source priority graphs, and quantified uncertainty --- covering structural variation that surface-level 
rephrasing cannot capture.
\textit{Third}, fine-grained configuration parameters across both default and inheritance reasoning (chain depth, branching factor, exception density, source count, hierarchy depth, sibling structure) make controlled diagnostic ablation a first-class capability. The implementation details are in Appendix~\ref{appendix:derelab_framework}.
\subsection{Defeasibility: Scope and Induction}

Defeasible reasoning is inherently \emph{non-monotonic}: conclusions warranted
by the current information may need to be retracted when new evidence arrives.
\textsc{DeReLab} operationalizes this directly by presenting information
incrementally rather than all at once. Each conversation begins with a partially
specified reasoning structure and then reveals additional facts one update at a
time. As the premise set evolves, previously supported conclusions may be
reinforced, defeated, reinstated, or left unchanged, depending on the structural
role of the new information.

This incremental construction induces a dynamic \emph{belief trajectory} over
the course of the conversation. An ideal defeasible reasoner should revise its
belief state precisely when new evidence changes the underlying inferential
structure, while remaining stable under irrelevant or out-of-scope updates.

\subsection{Reasoning Types}

\textsc{DeReLab} implements the two canonical paradigms of non-monotonic reasoning identified in the formal 
literature~\citep{10.5555/73682.73696}, each targeting a structurally distinct mode of defeasible inference.

\paragraph{Default Reasoning.}
Default reasoning concerns general rules that hold in the absence of specific contradicting information~\citep{REITER198081}: conclusions are drawn by default and retracted when exceptions arise.
In \textsc{DeReLab}, this is instantiated as a parameterized chain of default rules over objects, with updates that introduce confirmations, defeaters, distractors, and source-priority conflicts.

\paragraph{Inheritance Reasoning.}
Inheritance reasoning concerns the propagation of properties through a taxonomic hierarchy~\citep{10.5555/73682.73696, HORTY1990311}: a property attributed to a class is inherited by its subclasses and instances 
unless explicitly blocked at some node.
In \textsc{DeReLab}, this is instantiated as a configurable taxonomy with both \textbf{linear} and \textbf{tree} structures, where updates introduce inheritance-blocking exceptions or properties of sibling branches that must be correctly scoped to the hypothesis subject.




\subsection{Default Reasoning: Configuration and Question Types}

\paragraph{Configuration.}
We generate default reasoning examples through parameterized graphs controlled by chain length, number of object instantiations, distractor density, and a sparsity factor controlling defeat-edge injection. Two difficulty tiers are defined: \textbf{Easy} (shorter chains, fewer objects) and \textbf{Hard} (longer chains, more objects and distractors), with full parameter ranges in 
Appendix~\ref{appendix:derelab_framework_graph_gen}.
The order in which graph edges are revealed across turns is separately configurable, allowing the belief trajectory to be shaped by design: for instance, placing a defeating update early tests whether a model correctly reinstates a conclusion when subsequent confirming evidence arrives.

\paragraph{Question Types.}
Drawing on the taxonomy of default reasoning patterns in 
\citet{10.5555/73682.73696}, we identify three structural 
question types that cover the principal modes of defeasible 
inference; Appendix~\ref{appendix:default_reasoning} details how this 
taxonomy maps onto our graph structures.
\begin{itemize}
  \item \textbf{Type~1---Direct chain manipulation.}
    Positive or negative assertions about related properties test basic belief updating; exceptions test defeat of default rules.

  \item \textbf{Type~2---Irrelevant information.}
    An update concerns a property outside the default chain, making this variant irrelevant to the original default rules. This category checks the reasoning ability of isolating irrelevant property information.

  \item \textbf{Type~3---Source priority.}
    Competing claims from sources of differing reliability test whether models apply priority relations correctly, including when a higher-priority rule bypasses rather than directly 
    overrides a prior defeat.
\end{itemize}

In Figure \ref{fig:default_example}, a default reasoning example generation procedure is shown in detail.

\subsection{Inheritance Reasoning: Configuration and Question Types}

Following \citet{10.5555/73682.73696}, we distinguish two structural variants of inheritance reasoning and explicitly considering them as two different categories from here. \emph{Linear inheritance} arranges categories in a single chain from root to entity; \emph{tree inheritance} extends this with 
branching, where multiple sibling subcategories share a parent. The key diagnostic introduced by branching is scope: an update about a sibling branch must not affect the hypothesis subject, 
which belongs to a different branch entirely.

\paragraph{Configuration.}
Both variants are parameterized by hierarchy depth; tree inheritance is additionally parameterized by branching factor, controlling how many sibling subcategories share each parent node. Blocking edges are scoped to individual nodes: blocking inheritance at a node severs the chain for that node only, 
leaving all sibling and cousin nodes unaffected. The model must therefore track whether the hypothesis subject's specific path to the root carries an active block, rather than whether any path in the graph does. Full parameter ranges appear in Appendix~\ref{appendix:derelab_framework_graph_gen}.

\paragraph{Question Types.}
Unlike default reasoning, inheritance conversations pose only the entailment question (\textit{yes\,/\,no\,/\,unknown}) after each update. The belief-update question is omitted because the direction of change is directly recoverable from consecutive entailment answers: a transition from \textit{yes} to \textit{no} is by definition a weakening, making a separate effect question redundant. The primary diagnostic is therefore whether the model correctly identifies which updates fall within the scope of the hypothesis subject's inheritance path and which do not.

In Figures \ref{fig:linear_example} and \ref{fig:tree_example}, the generation procedure for two inheritance reasoning examples (linear and tree) is shown in detail.




\subsection{Graph Resolution}
\label{sec:resolution}

Ground truth labels in \textsc{DeReLab} are produced deterministically by a path-based resolution procedure grounded in the inheritance theory of \citet{HORTY1990311}. Given the accumulated premise set at any conversational turn, the resolver traverses the knowledge graph and returns one of three verdicts for the target hypothesis: \textsc{Entailed} (\textit{yes}), \textsc{Defeated} (\textit{no}), or \textsc{Undetermined} (\textit{unknown}). This mechanism produces formally correct, annotation-free labels at every conversational turn when new information is revealed. The full resolution procedure is described in Appendix~\ref{appendix:path_resolver}.

\subsection{Conversation Structure}

Across all categories, each conversation contains:
(i)~a \emph{pretext section} presenting background default rules and entity
instantiations;
(ii)~a \emph{hypothesis} stating the entailment question under evaluation;
and (iii)~a sequence of \emph{update turns}, each delivering one new fact
followed by the entailment question.
For default reasoning, each update turn is followed by a second \emph{effect question} eliciting whether the update strengthened, weakened, or had no effect on hypothesis support. Default reasoning conversations may chain multiple hypothesis sections via a new hypothesis turn that changes the active subject while preserving the full set of accumulated premises. Furthermore, default reasoning can have multiple competing sources with priority rankings, from which an update originates, where a higher-priority source will supersede an update from a lower-priority one. For our experimental purpose, we construct an alternate version where the effect question comes prior to the entailment question. (see Appendices \ref{appendix:data_samples}  and \ref{appendix:reversed_metacog} for details)

\begin{table*}[t]
\centering
\footnotesize
\setlength{\tabcolsep}{4pt}
\begin{tabular}{clp{2cm}p{4.5cm}p{6.5cm}}
\toprule
\textbf{Cond.} & 
\textbf{Prior} & 
\textbf{GT effect} & 
\textbf{Meaning} & 
\textbf{What it tests} \\
\midrule
C1 & \textit{yes} & strengthening
  & Model believes hypothesis; new evidence supports it
  & Whether models correctly accept congruent evidence
    (control condition) \\[3pt]
C2 & \textit{yes} & weakening
  & Model believes hypothesis; new evidence defeats it
  & Whether models revise beliefs against their prior
    conclusion --- the primary confirmation bias signal \\[3pt]
C3 & \textit{no}  & no effect (neg.)
  & Hypothesis already defeated; negative-sounding update arrives
  & Whether models react to surface negation rather than
    logical scope (\emph{semantic override}) \\[3pt]
C4 & \textit{no}  & no effect (neutral)
  & Hypothesis already defeated; neutral update arrives
  & Whether a logically irrelevant distractor
    causes spurious belief revision (neutral baseline) \\[3pt]
C5 & \textit{unk.} & str.\,/\,weak.
  & Model is uncertain; directional evidence arrives
  & Whether models correctly resolve from an indeterminate
    state (\emph{anchoring under uncertainty}) \\
\bottomrule
\end{tabular}
\caption{Congruence conditions assigned to each update turn.
\textbf{Prior} is the model's own predicted answer at
the immediately preceding turn, not the ground truth.
\textbf{GT effect} is the formally verified effect of the
incoming update on the hypothesis.
Confirmation bias is evidenced by a systematic accuracy gap
between C2 and C1 turns within the same conversation.}
\label{tab:congruence}
\end{table*}

\subsection{Entity Design and Contamination Prevention}

The overlap between benchmark content and pretraining corpora is a recognized threat to the validity of LLM evaluation ~\citep{magar-schwartz-2022-data, allawati2026llmbenchmarkdatasetscontaminationresistant, zhou2023dontmakellmevaluation}. Although our proposed framework is flexible to use any type of entity, we present results keeping all entities in the primary \textsc{DeReLab} dataset as \emph{pseudowords} (logatomes) pronounceable but semantically null strings generated programmatically
(e.g.,\ \textit{Kitylu}, \textit{Atool}, \textit{Deraw}) (see Appendix \ref{appendix:data_creation_nonce} for pseudoword generation details). For defeasible reasoning, a model may produce the correct answer by retrieving a memorized real-world fact (\textit{penguins cannot fly}) rather than by executing the defeasible inference the benchmark intends to measure. Because pseudoword entities carry no prior semantics in any trained model, correct responses can only arise from reasoning over the premises supplied in context.

\subsection{Cognitive Bias Probing}
\label{sec:cogsci_framework}

A key capability of \textsc{DeReLab} is the ability to embed
cognitive science-inspired experiments directly into the evaluation
structure. Because the framework controls the content of information updates, it can construct controlled experimental conditions.

As a primary instantiation, we measure \emph{confirmation bias} in defeasible belief updating: the tendency of LLMs to accept evidence congruent with their current belief while
resisting contradicting evidence~\citep{nickerson1998confirmation}.
Each update turn is assigned a congruence condition based on the model's own prior predicted answer and the ground truth effect of the update. Table \ref{tab:congruence} defines the congruence conditions. 

We use two complementary metrics to measure confirmation bias. The \textbf{Bias Gap} measures the accuracy difference 
between congruent and incongruent conditions:
\begin{equation}
\text{BiasGap} = \text{Acc}(\text{C1}) - \text{Acc}(\text{C2})
\end{equation}
A positive BiasGap indicates the model handles supporting evidence more accurately than defeating evidence --- 
the hallmark of confirmation bias. The \textbf{Odds Ratio} (OR) provides a scale-invariant effect size:
\begin{equation}
\text{OR} = \frac{\text{C1}_{\text{correct}} 
\times \text{C2}_{\text{wrong}}}
{\text{C1}_{\text{wrong}} \times 
\text{C2}_{\text{correct}}}
\end{equation}
OR\,$> 1$ indicates confirmation bias; OR\,$= 1$ 
indicates no asymmetry between conditions.
Bootstrap 95\,\% confidence intervals are computed at 
the conversation level.
Full operationalization details appear in Appendix~\ref{appendix:confirmation_bias}.

\begin{figure*}[h]
    \centering

        \centering
        \includegraphics[width=\linewidth]{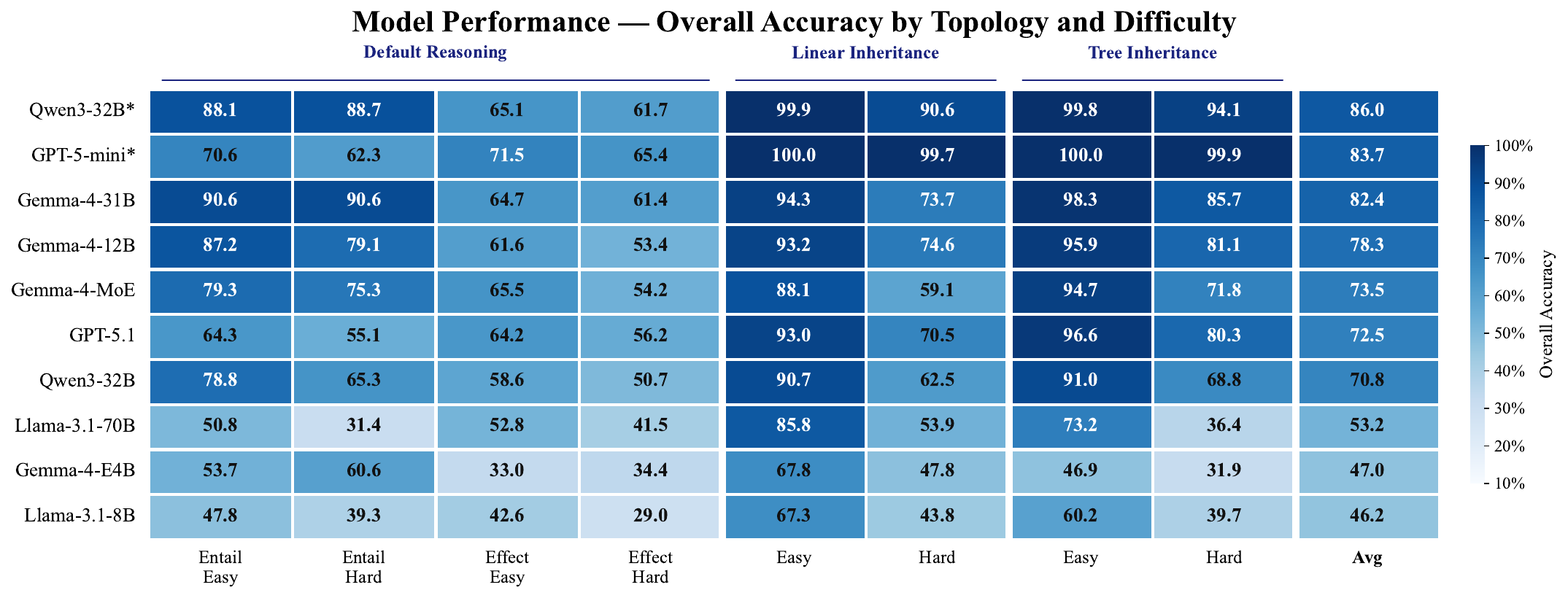}
        \label{fig:performance_heatmap}

    \caption{Turn-level model performance across reasoning topologies and difficulty levels. Each cell reports accuracy for a model under one topology--difficulty condition. (*) represents reasoning-enabled inference.}
    \label{fig:model_performance}
\end{figure*}

\section{Experimental Setup}
\label{sec:experimental_setup}

\subsection{Dataset}
\label{sec:dataset}
\begin{table}[h]
\centering
\resizebox{\linewidth}{!}{
\begin{tabular}{llrr}
\toprule
Topology &
\begin{tabular}[c]{@{}l@{}}Difficulty\\ Configuration\end{tabular} &
\begin{tabular}[c]{@{}r@{}}Conver-\\ sations\end{tabular} &
\begin{tabular}[c]{@{}r@{}}Question\\ Turns\end{tabular} \\
\midrule
Default reasoning & Easy & 150 & 1,905 \\
Default reasoning & Hard & 150 & 9,325 \\
Linear inheritance & Easy & 150 & 646 \\
Linear inheritance & Hard & 150 & 2,483 \\
Tree inheritance & Easy & 150 & 963 \\
Tree inheritance & Hard & 150 & 3,068 \\
\midrule
Total & -- & 900 & 18,390 \\
\bottomrule
\end{tabular}
}
\caption{Canonical \textsc{DeReLab} evaluation set. ``Questions'' refers to the
total number of answer-bearing turns evaluated across all conversations in each
split.}
\label{tab:dataset}
\end{table}
We evaluate models on three \textsc{DeReLab} reasoning topologies: default reasoning, linear inheritance, and tree inheritance. For each topology,
we generate easy and hard splits, with 150 conversations per topology-difficulty combination. This gives 900 canonical evaluation conversations in total (Table \ref{tab:dataset}). All canonical examples use
pseudoword entities to avoid contamination from memorized real-world facts. The easy and hard splits differ in graph size, distractor attributes, density, and reasoning depth (see Table \ref{tab:inherit_difficulty_presets} and \ref{tab:parameters} in Appendix \ref{appendix:derelab_framework}). 
We additionally construct a robustness dataset: a reseeding set of 1,200 examples to test whether results are stable across random seeds. This set contains 20 seeds,
2 topologies, and 30 conversations per topology per seed.

\subsection{Models}

\begin{table}[h]
\centering
\resizebox{\linewidth}{!}{
\begin{tabular}{ll}
\toprule
Model & Type \\
\midrule
\texttt{gpt-5.1}
& Closed, standard \\

\texttt{gpt-5-mini}
& Closed, compact reasoning \\

\texttt{Gemma-4-31B-it}
& Open, instruction-tuned \\

\texttt{Gemma-4-26B-A4B}
& Open, Mixture of Expert (MoE) \\

\texttt{Gemma-4-12B-it}
& Open, instruction-tuned \\

\texttt{Gemma-4-E4B}
& Open, instruction-tuned \\

\texttt{Llama-3.1-70B}
& Open, instruction-tuned \\

\texttt{Llama-3.1-8B}
& Open, instruction-tuned \\

\texttt{Qwen3-32B (*)}
& Open, non-thinking and thinking\\

\bottomrule
\end{tabular}
}
\caption{Models evaluated in our experiments. (*) We use Qwen3-32B model in both thinking and non-thinking mode for inference. Further details about the models are documented in Appendix \ref{appendix:resources}.}
\label{tab:models}
\end{table}

We evaluate both closed and open-weight models, covering different model
families, sizes, and architectures. The closed GPT models provide a strong
frontier-model reference point. The Llama and Gemma-4 models provide open-weight
comparisons across model families and scales. For architectural diversity, we also include a mixture-of-experts (MoE) model from the Gemma 4 (\texttt{Gemma-4-26B-A4B}, hereafter referred to as \texttt{Gemma-4-MoE}) family alongside dense models. Furthermore, for direct comparison between thinking and non-thinking modes, we utilize the dual capability of Qwen3-32B model (Table \ref{tab:models}). We evaluate models in a multi-turn chat setting.

\begin{figure*}[t]
    \centering

    \begin{subfigure}[t]{0.49\linewidth}
        \centering
        \includegraphics[width=\linewidth]{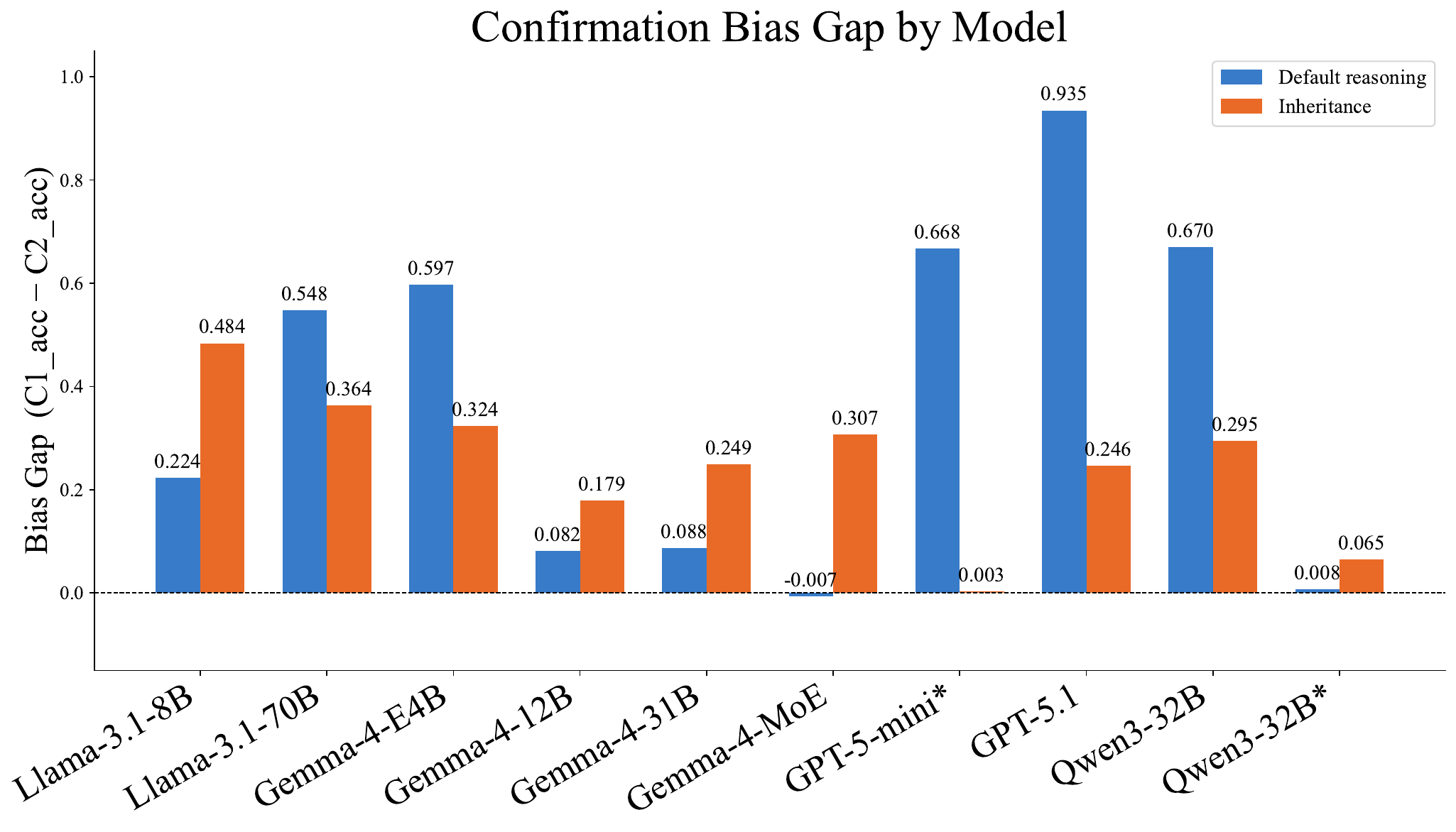}
        \caption{Confirmation bias gap (\textbf{C1\,acc $-$ C2\,acc}) per model
and reasoning type. Gemma-4-MoE on default reasoning and Qwen3-32B* on inheritance are non-significant after Holm correction.
OR undefined for GPT-5-mini inheritance (C1 errors\,$= 0$).}
        \label{fig:bias_gap_by_model}
    \end{subfigure}
    \hfill
    \begin{subfigure}[t]{0.49\linewidth}
        \centering
        \includegraphics[width=\linewidth]{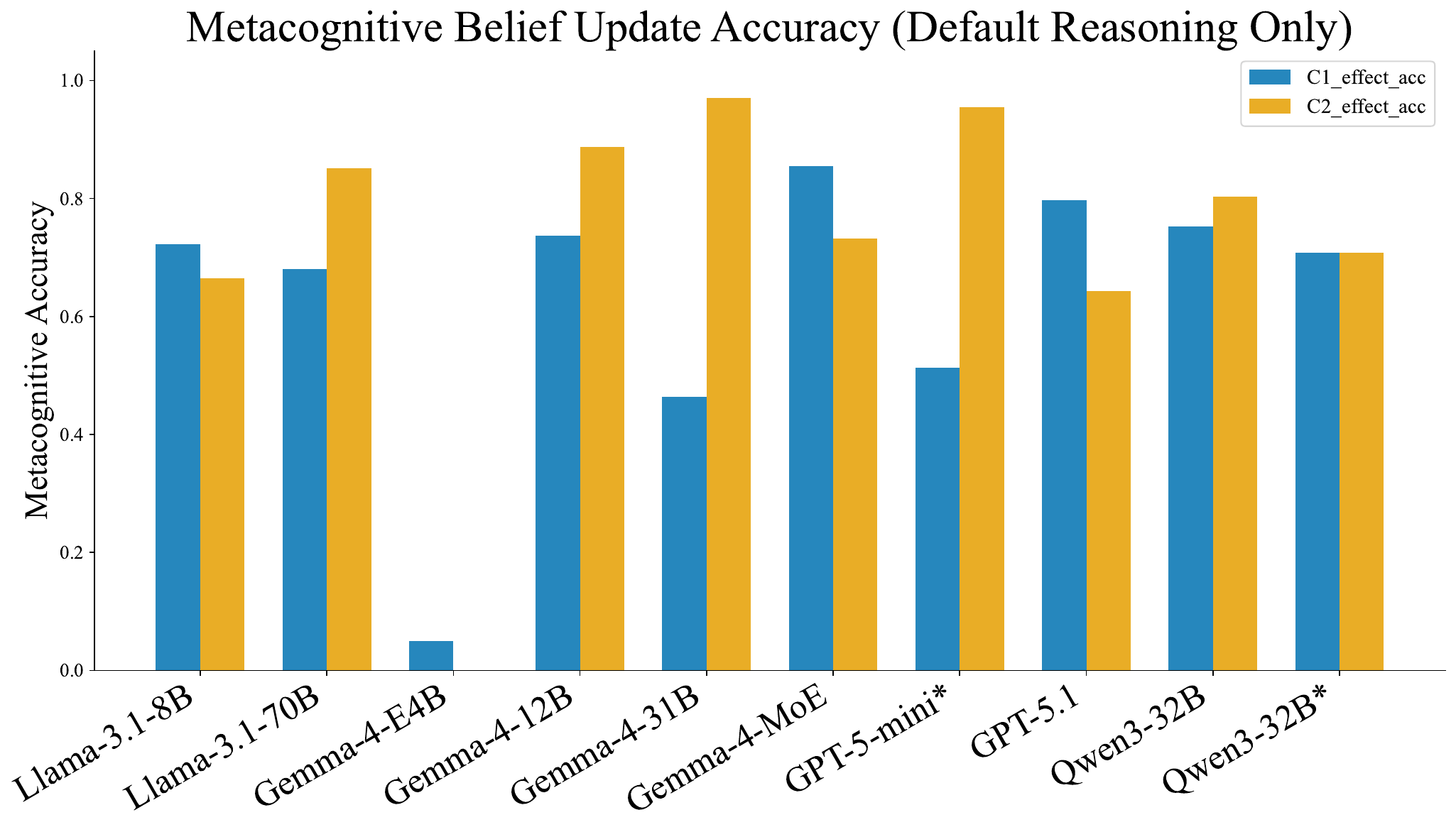}
        \caption{Metacognitive accuracy on the belief-update task
(default reasoning only).
\textbf{C1-effect}: fraction of congruent turns correctly
labeled \textit{strengthening}.
\textbf{C2-effect}: fraction of incongruent turns correctly
labeled \textit{weakening}.}
        \label{fig:matabars}
    \end{subfigure}

    \caption{Confirmation bias magnitude and metacognitive awareness across models.
\textbf{Left}: BiasGap (C1\,acc $-$ C2\,acc) measures how much more
accurately a model handles congruent than incongruent updates, broken
down by reasoning type.
\textbf{Right}: Accuracy on the separate effect-of-update question
for the same turns, distinguishing whether models that fail the
entailment question nonetheless correctly identify the direction of
the update.}
    \label{fig:confirmation_metabar_biasgap_effect}
\end{figure*}

\section{Results and Discussion}
\label{sec:results}
In the following sections, we analyze our findings about model performance and behavior using data generated through \derelab{}. 

\subsection{Overall Reasoning Accuracy}
\label{sec:overall_accuracy}

\textit{How capable are current LLMs at defeasible reasoning, and does capability vary systematically across reasoning type and task difficulty?}

We evaluate all the models on the generated dataset for both default and inheritance reasoning at easy and hard difficulty levels. We report turn-level accuracy against ground-truth labels and performance with respect to reasoning complexity in Figure \ref{fig:model_performance}. 

We notice a significant performance gap between reasoning-enabled and standard instruction-tuned configurations, where reasoning models clearly outperform the latter. Within task type, default reasoning is consistently harder than inheritance reasoning, and the belief-update task (strengthening/weakening/no effect) is the most demanding of all conditions. Larger models perform better within each family: Gemma-4-31B > Gemma-4-12B > Gemma-4-E4B and Llama-3.1-70B > Llama-3.1-8B across all conditions, consistent with the general scaling trend observed in reasoning benchmarks. For Qwen3-32B, performance decreases substantially when extended thinking is disabled.

\subsection{Confirmation Bias in Belief Updating}
\label{sec:confirmation_bias_results}
\textit{Do LLMs exhibit confirmation bias when updating beliefs under
incongruent evidence, and what does this reveal about the nature of
their belief revision?}

We apply the congruence-condition framework (\S\ref{sec:cogsci_framework})
to all Phase~1 inference results, yielding between 9,029 and 11,682
labeled turns per model across all reasoning types and difficulty tiers.
Statistical methodology and full supplementary results appear in
Appendix~\ref{appendix:confirmation_bias} and \ref{app:cogsci_results} respectively.

\paragraph{Confirmation bias is near-universal.}
Confirmation bias is near-universal: 17 of 19 model--reasoning-type
cells are statistically significant after Holm--Bonferroni correction
(Table~\ref{tab:bias_main}, Figure~\ref{fig:bias_gap_by_model}).
Bias magnitude spans a wide range: Gemma-4-12B shows a small but
reliable gap (BiasGap\,=\,0.082, OR\,=\,2.9), while GPT-5.1 reaches
near-total incongruent failure (BiasGap\,=\,0.935, OR\,=\,891,
95\,\% CI\,[586,\,1498]) --- a C2 accuracy of just 3.5\,\% means
the model almost never correctly revises from \textit{yes} to
\textit{no} when faced with a weakening update.
Two cells are non-significant after correction: Gemma-4-MoE default
(Gap\,=\,$-$0.007) and Qwen3-32B* inheritance (Gap\,=\,0.065).
As we show below, the Gemma-4-MoE result is an artifact of
question ordering rather than a genuine absence of bias.

\begin{table}[t]
\centering
\small
\setlength{\tabcolsep}{2.5pt}
\begin{tabular}{l ccc ccc}
\toprule
&
\multicolumn{3}{c}{\textbf{Default}} &
\multicolumn{3}{c}{\textbf{Inheritance}} \\
\cmidrule(lr){2-4}
\cmidrule(lr){5-7}
\textbf{Model}
& \textbf{Gap} & \textbf{OR} & \textbf{Anch.}
& \textbf{Gap} & \textbf{OR} & \textbf{Anch.} \\
\midrule
GPT-5.1        & .94 & 891  & .43 & .25 &  9.3 & .05 \\
GPT-5-mini     & .67 & 26.5 & .02 & .00 &  --- & .00 \\
Qwen3-32B & .67 & 93.3 & .48 & .30 &  6.8 & .22 \\
Gemma-4-E4B    & .60 & 72.7 & .40 & .32 &  4.0 & .64 \\
Llama-3.1-70B  & .55 & 13.2 & .19 & .36 &  7.8 & .07 \\
Llama-3.1-8B   & .22 &  2.8 & .12 & .48 &  8.3 & .34 \\
Gemma-4-31B    & .09 & 18.1 & .05 & .25 &  6.6 & .05 \\
Gemma-4-12B    & .08 &  2.9 & .12 & .18 &  2.8 & .14 \\
Qwen3-32B*     & .01 &  1.1 & .03 & .07 &  3.7 & .00 \\
Gemma-4-MoE    &$-.01$& 0.8 & .03 & .31 & 16.7 & .02 \\
\bottomrule
\end{tabular}
\caption{
Confirmation-bias metrics across default and inheritance reasoning.
\textbf{Gap} = C1\,acc $-$ C2\,acc.
\textbf{OR} = odds ratio.
\textbf{Anch.} = anchoring rate on ground-truth-flipping sections.
Qwen3-32B* denotes the extended-thinking variant.
OR undefined (---) when C1 contains zero errors.
}
\label{tab:bias_main}
\end{table}

\paragraph{Bias is stronger on default reasoning than inheritance.}
For most models, the BiasGap is larger on default reasoning than
inheritance (Table~\ref{tab:bias_main}).
The contrast is sharpest for GPT-5-mini (0.668\,$\to$\,0.003) and
GPT-5.1 (0.935\,$\to$\,0.246).
We attribute this to the richer update vocabulary of default
reasoning --- source-priority conflicts, trap distractors, and
multi-turn exception chains create more opportunities for prior-belief
anchoring than the structurally simpler inheritance updates.
Two exceptions to this phenomenon are Llama-3.1-8B (default 0.224, inheritance 0.484)
and Gemma-4-MoE (default $-$0.007, inheritance 0.307).

\paragraph{Metacognitive dissociation:}
Each update turn contains two distinct questions:
the \emph{entailment question} (reported as C1\,acc and C2\,acc) and the
\emph{effect-of-update question}
(\textit{strengthening\,/\,weakening\,/\,no-effect},
reported as C1-eff and C2-eff).
A pattern emerges when we examine them together
(Figure~\ref{fig:confirmation_metabar_biasgap_effect}): several models correctly label
a C2 update as \textit{weakening} at high rates yet simultaneously
answer \textit{yes} to the entailment question on the same turn.
The largest absolute dissociation are in GPT-5-mini
(C2-eff\,=\,95.4\,\%, C2\,acc\,=\,20.3\,\%) and Llama-3.1-70B
(C2-eff\,=\,85.1\,\%, C2\,acc\,=\,14.9\,\%): both models correctly
identify weakening on the majority of C2 turns yet almost never
revise their entailment answer accordingly.
We refer to this discrepancy as a \emph{know-but-don't-output} pattern: the model produces the correct effect label but an inconsistent entailment judgment on the same turn.

\paragraph{Reversed question order reveals scaffolded causal labeling.}
We reverse the order of the entailment and effect questions to test whether high C2-eff reflects independent causal understanding or is scaffolded by prior engagement
with the entailment judgment. C2-eff dropped substantially across all tested models:
Gemma-4-31B (0.970\,$\to$\,0.264), Gemma-4-12B
(0.887\,$\to$\,0.209), Gemma-4-MoE (0.732\,$\to$\,0.288), and
Qwen3-32B (0.803\,$\to$\,0.414).
Gemma-4-E4B is the lone exception (0.000\,$\to$\,0.349): its
near-zero original score reflected a vocabulary failure rather than
a dissociation.
Entailment accuracy on C2 turns also worsened under reversal for
most models, with the BiasGap rising substantially ---
most strikingly for Qwen3-32B (0.670\,$\to$\,0.816) and
Gemma-4-MoE ($-$0.007\,$\to$\,0.259, exposing the near-zero default
gap as an ordering artifact).
Together, these results show the entailment question acts as a broad
cognitive scaffold: engaging with the yes/no/unknown judgment first
triggers reasoning that benefits \emph{both} the effect label and
the entailment answer itself.
The know-but-don't-output dissociation observed in the standard order
is therefore a weaker form of the same effect --- the scaffold is
strong enough to enable correct effect labeling but insufficient to
overcome the entailment bias on most models.
Full comparison tables appear in Appendix~\ref{appendix:reversed_metacog}.

\paragraph{Thinking mode suppresses bias.}
Qwen3-32B* (Qwen3-32B with extended thinking) achieves near-zero
default bias in the standard evaluation order
(Gap\,=\,0.008, OR\,=\,1.07; Holm\,$p = 0.015$ but effect negligible)
and the lowest anchoring rates of any model on both topologies
(default: 0.032; inheritance: 0.004).
The standard Qwen3-32B, by contrast, shows a default BiasGap of 0.670
--- one of the largest in the evaluation, and its gap rises further
to 0.816 under reversed question order. This suggests that the thinking scaffold has a substantial effect on both the entailment and effect questions, which could be attributed to figuring out the structure of the graph over which the information is presented. 

\paragraph{Anchoring and semantic override.}
At the conversation level, GPT-5.1 and Qwen3-32B exhibit the highest
default anchoring rates (0.433 and 0.483): in nearly half of
conversations where the ground-truth answer changed, these models
maintained their initial prediction.
Gemma-4-E4B shows the most persistent anchoring on inheritance
(0.635), consistent with its strong prior-preserving behavior
across both metrics.
By contrast, Gemma-4-31B and GPT-5-mini show near-zero anchoring
($\leq$\,5\,\%), indicating per-turn bias does not accumulate into
global belief fixation for these models.
Llama-3.1-70B shows the highest semantic override rate on default
(83.9\,\%): it mislabels C3 turns as \textit{weakening} even when
the chain is already defeated, reflecting strong sensitivity to
surface-level negative language regardless of logical scope.
Full results are in Appendix~\ref{appendix:anchoring_override}.

\subsection{Confidence and Inference Path Length}
\label{sec:path_confidence}
\textit{Does model confidence in the correct answer decay as the
inference chain between the updated node and the hypothesis grows
longer?}

We analyze tree-inheritance turns along two dimensions: \emph{on-path} turns, where the update is causally relevant to the hypothesis, and \emph{off-path} turns, where the update concerns a node outside the hypothesis subject's inheritance path and should have no effect on confidence. We measure $\Delta\text{gold\_conf}$ --- the change in probability assigned to the correct label relative to the model's initial answer --- as our primary confidence signal. Full statistical details appear in Appendix~\ref{appendix:path_confidence}.

\begin{table}[t]
\centering
\footnotesize
\setlength{\tabcolsep}{3pt}
\resizebox{\columnwidth}{!}{%
\begin{tabular}{l c cc}
\toprule
& \textbf{On-path} $\rho$ &
  \multicolumn{2}{c}{\textbf{Off-path} $\rho$} \\
\cmidrule(lr){2-2}\cmidrule(lr){3-4}
\textbf{Model} & Hard & Easy & Hard \\
\midrule
GPT-5.1       & $+.13^{*}$   & $+.09^{*}$  & $+.01$      \\
Qwen3-32B     & $-.03$       & $+.06$      & $+.01$      \\
Gemma-4-12B   & $-.16^{**}$  & $-.01$      & $+.01$      \\
Gemma-4-MoE   & $-.18^{**}$  & $-.12^{**}$ & $-.04$      \\
Qwen3-32B*    & $-.01$       & $-.13^{***}$& $-.03$      \\
Gemma-4-31B   & $-.30^{***}$ & $-.01$      & $-.03$      \\
Llama-3.1-8B  & $-.26^{***}$ & $-.06$      & $-.01$      \\
Llama-3.1-70B & $-.21^{***}$ & $+.02$      & $-.03$      \\
Gemma-4-E4B   & $+.07$       & $+.01$      & $+.10^{***}$\\
\bottomrule
\end{tabular}}
\caption{Spearman $\rho$ between update-to-hypothesis distance and
correct-answer confidence.
\textbf{On-path}: hard tree-inheritance turns only; easy omitted
because of insufficient distance variation.
\textbf{Off-path}: model's response to causally irrelevant updates;
values near zero indicate correct suppression.
Negative on-path $\rho$ = confidence decays with distance;
positive = accumulation.
$^{*}p{<}.05$, $^{**}p{<}.01$, $^{***}p{<}.001$.}
\label{tab:path_spearman}
\end{table}

On-path, correct-answer confidence declines significantly with
distance for most models (Table~\ref{tab:path_spearman}) --- most
steeply for Gemma-4-31B ($\rho = -0.298$, $p < 0.001$),
Llama-3.1-8B ($\rho = -0.255$, $p < 0.001$), and
Llama-3.1-70B ($\rho = -0.207$, $p < 0.001$) --- while GPT-5.1
shows the opposite ($\rho = +0.130$, $p = 0.026$), consistent with
successfully accumulating evidence over longer chains.
Off-path, Llama-3.1-8B changes its answer incorrectly on more than
one in three irrelevant turns (36\,\%), whereas GPT-5.1 and
Gemma-4-31B do so on fewer than 1\,\%: the models that maintain
confidence along longer chains are also those that correctly ignore
updates that should not affect them.


\begin{figure}[t]
    \centering

    \begin{subfigure}{0.8\linewidth}
        \centering
        \includegraphics[width=\linewidth]{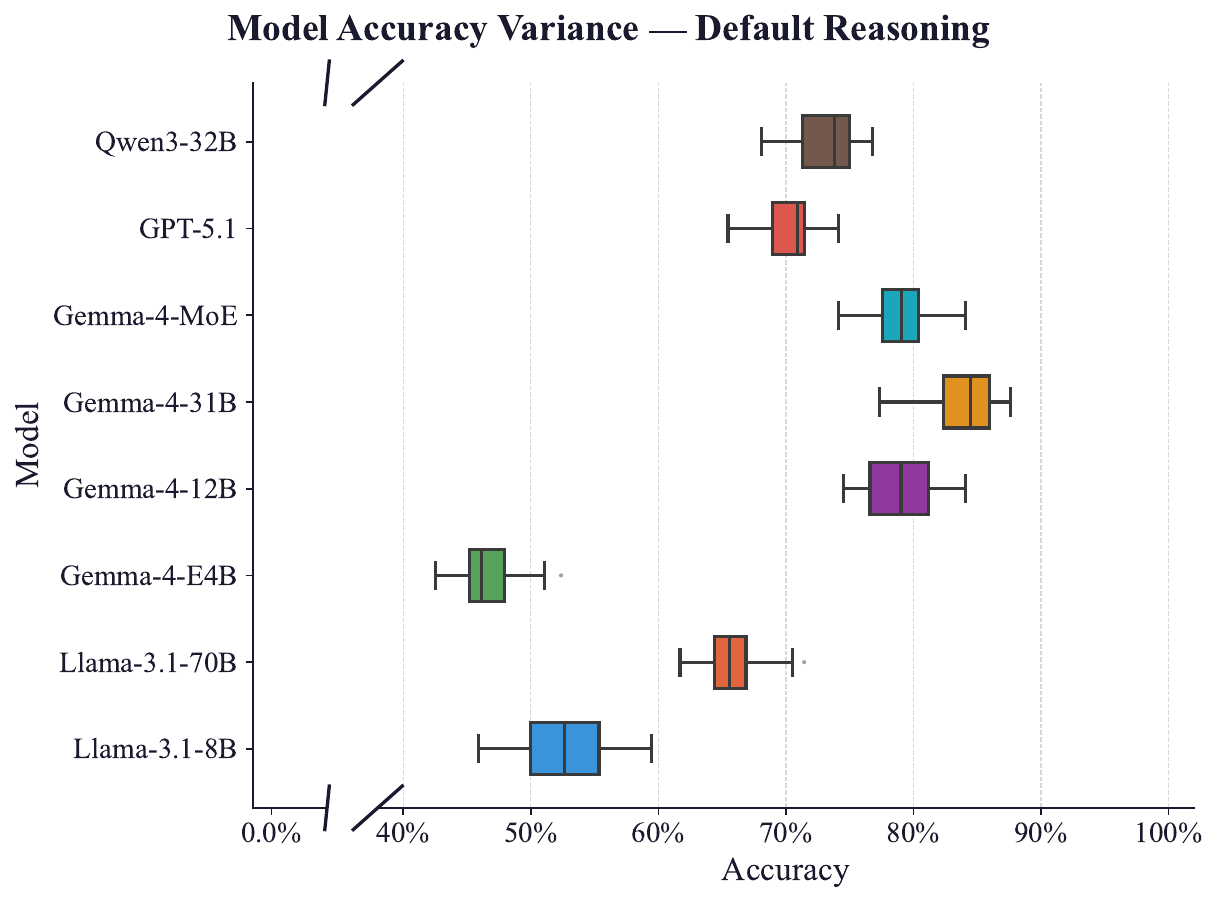}
        \caption{Variance in accuracy by property chain length category for default reasoning.}
        \label{fig:reseed_accuracy}
    \end{subfigure}

    \begin{subfigure}{0.8\linewidth}
        \centering
        \includegraphics[width=\linewidth]{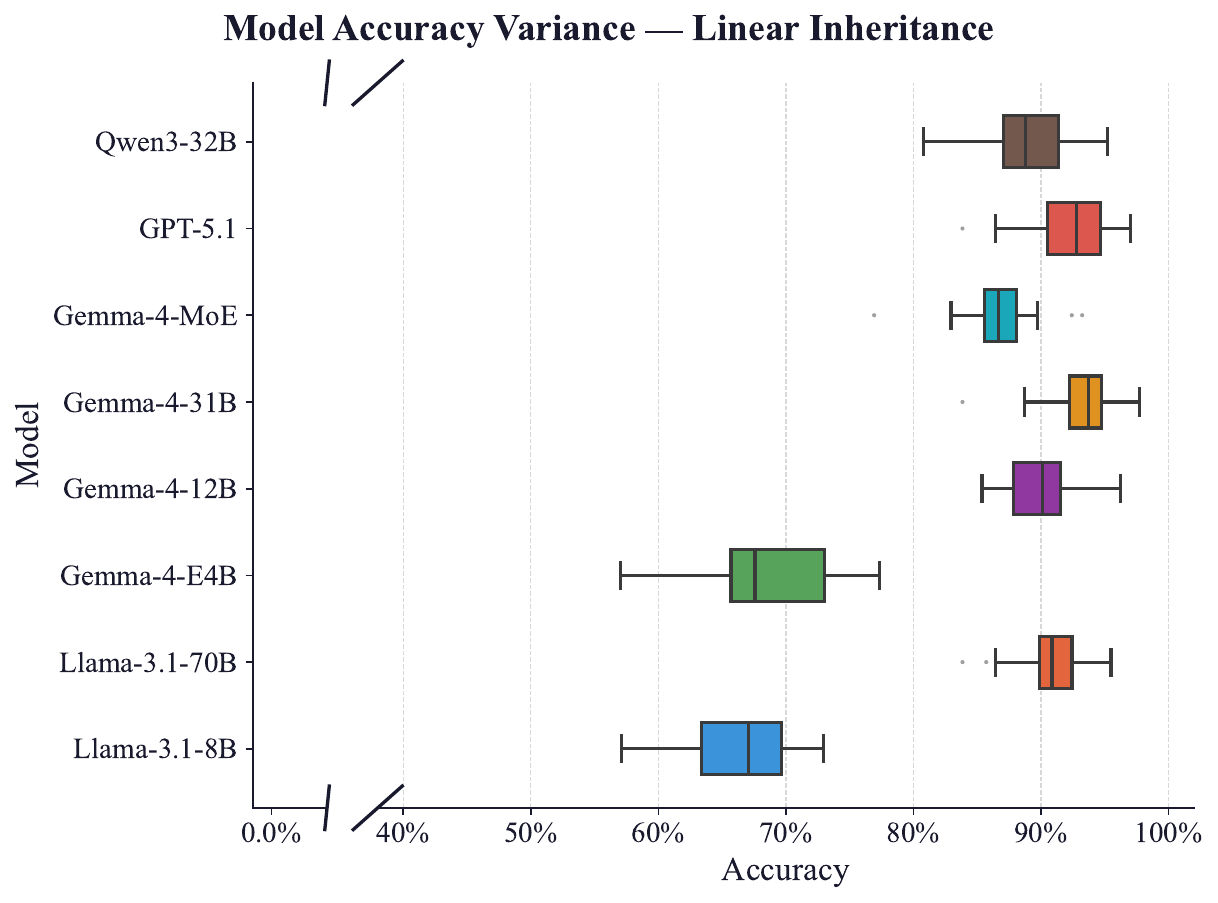}
        \caption{Variance in accuracy by inheritance hierarchy depth for inheritance reasoning (linear).}
        \label{fig:reseed_stability}
    \end{subfigure}

    \caption{Reseeding robustness analysis.}
    \label{fig:reseed_results}
\end{figure}

\subsection{Seeded Sample Variance Analysis}
\label{sec:seeded_variance}

\textit{Are \textsc{DeReLab} evaluations reproducible across 
independent random seeds, and does stability hold across 
model tiers?}

We evaluate eight models on 20 independently seeded draws of the evaluation data (default and linear inheritance, easy difficulty) and measure variability. Three statistics are reported: the intraclass correlation coefficient
(ICC, two-way random effects, absolute agreement), Kruskal--Wallis tests
for per-sample accuracy distributions, and the coefficient of
variation (CV\,$=$\,SD\,/\,mean\,$\times$\,100).
Figure~\ref{fig:reseed_results} shows the resulting boxplots; 
full statistical details appear in 
Appendix~\ref{appendix:stat_sig_reseed}.

\paragraph{Scores are highly reproducible across seeds.}
The ICC across all model--condition cells is 0.949
(95\,\% CI [0.938, 0.963]), indicating that over 94\,\% of the
variance in turn-level accuracy is shared across seeded draws, confirming that the benchmark captures consistent model behavior.
Kruskal--Wallis tests yield no significant differences across the 20
seeds for any model--condition pair (all $p > 0.06$, after
accounting for multiple comparisons), confirming that seed choice
does not systematically alter the ranking or magnitude of scores.

Our results also show that within-model variance is low for most models. The CV is below 4\,\% for five out of eight models on both default and linear-inheritance conditions. (Appendix \ref{appendix:stat_sig_reseed}, Table \ref{tab:kruskal_wallis}). These results confirm that the benchmark scores reported throughout this paper reflect stable model properties rather than sampling variability.


\section{Conclusion}
We introduced \derelab{}, a structural framework for evaluating
defeasible reasoning in LLMs through formally verified, multi-turn
benchmarks. Our experiments reveal that current models are brittle under
belief-updating conditions: confirmation bias is near-universal,
confidence degrades with inference-path length, and structural
relevance is rarely tracked reliably.
Future work can extend the framework with additional non-monotonic
reasoning structures and further cognitive science-inspired probes
to deepen understanding of where and why defeasible reasoning fails.

\newpage
\section*{Limitations}



The current framework evaluates the default and inheritance topologies, which can be further expanded to support a broader spectrum of defeasible reasoning structures.
Moreover, \derelab{} didn't evaluate models' consistency against human performance, leaving it an open question whether human reasoning diverges from algorithmic conclusions. 

Our evaluation includes only two explicitly reasoning-enabled configurations—GPT-5-mini and Qwen3-32B in extended-thinking mode. More models could help us find patterns. We found biases exist but didn't discuss why they were found.   

\bibliography{custom}

\appendix

\newpage

\section{\derelab~Framework}
\label{appendix:derelab_framework}
\begin{figure*}[h]
    \centering
    \includegraphics[width=\linewidth]{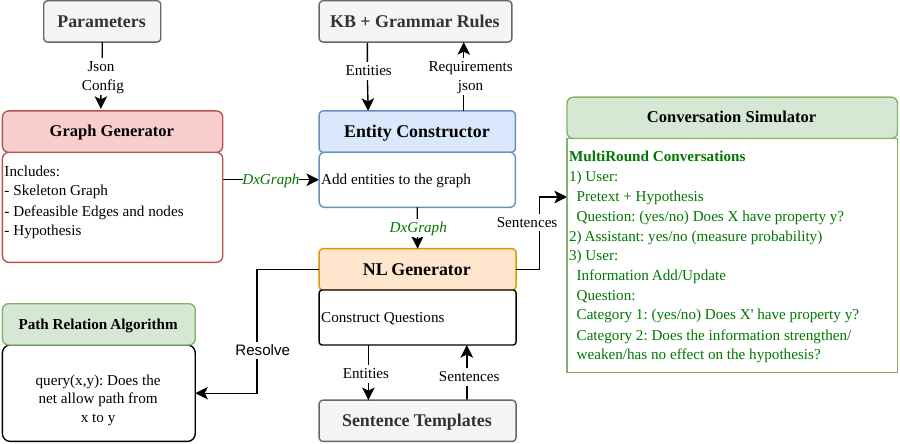}
    \caption{Modules of \derelab{} framework used for data generation.}
    \label{fig:appendix-modules}
\end{figure*}

We discuss the different modules of \derelab{} in the following sections. The logical flow is depicted in Figure \ref{fig:appendix-modules}.

\subsection{Graph Generation}
\label{appendix:derelab_framework_graph_gen}

Our reasoning structure contains two node types: entity and property nodes. The graph contains four primary edge types. These are: 

\begin{itemize}
    \item \textbf{\texttt{Hypothesis Edge}}: Represents the target query or tentative conclusion that is currently under evaluation. 

    \item Has Attribute: Denotes factual and monotonic assignment of a property to an entity.

    \item Defeasible Has Property: Represents a non-monotonic implication indicating that an entity ``generally'' or ``typically'' possesses a certain property. For example, ``Red objects are generally cubic''.

    \item Defeasible Lack Property: Represents a non-monotonic implication indicating that an entity ``generally'' does not possess a specific trait. For example, ``Cubic objects generally do not roll''.

\end{itemize}

The graph generator module could construct three distinct topologies:

\textbf{Linear Inheritance:} The graph formed a single, unbranched entity chain from a root concept down to a terminal subclass. The root concept was assigned a base target property. To introduce non-monotonicity, defeasible edges were probabilistically injected at intermediate nodes in the chain. The final hypothesis strictly queried the leaf node.

\textbf{Tree Inheritance:} The tree expanded from a root node where the maximum branching factor exponentially decayed with depth. Branches also underwent random survival checks, resulting in asymmetrical structures. Defeasible edges were randomly added across intermediate nodes, and the hypothesis edge targeted a leaf node.

In Table \ref{tab:inherit_difficulty_presets}, the parameters tuned to set the difficulty of Inheritance Reasoning topologies across different presets are listed: easy (fewer nodes, less depth and breadth) and hard.
\begin{table}[h]
    \centering
    \small

    \begin{tabular}{lcc}
        \toprule
        \textbf{Parameter} & \textbf{Easy} & \textbf{Hard} \\
        \midrule
        
        \multicolumn{3}{@{}l}{\textbf{Global Settings}} \\
        Sparsity factor range & 0.50--0.80 & 0.40--0.60 \\
        Serial random         & True       & True       \\
        \midrule
        
        \multicolumn{3}{@{}l}{\textbf{Jellyfish Preset}} \\
        Max depth            & 4    & 8      \\
        Root breadth         & 5    & 13     \\
        Branching (min--max) & 1--1 & 1--1   \\
        Survival probability & 0.40 & 0.60   \\
        Branching decay      & 1.0  & 1.0    \\
        Expected nodes       & 4--8 & 24--50 \\
        \midrule
        
        \multicolumn{3}{@{}l}{\textbf{Balanced Preset}} \\
        Max depth            & 3    & 4    \\
        Root breadth         & 2    & 3    \\
        Branching (min--max) & 2--2 & 2--2 \\
        Survival probability & 1.0  & 1.0  \\
        Branching decay      & 1.0  & 1.0  \\
        Expected nodes       & 7    & 46   \\
        \midrule
        
        \multicolumn{3}{@{}l}{\textbf{Bushy Preset}} \\
        Max depth            & 2    & 4      \\
        Root breadth         & 3    & 4      \\
        Branching (min--max) & 1--2 & 1--4   \\
        Survival probability & 0.70 & 0.80   \\
        Branching decay      & 1.0  & 0.75   \\
        Expected nodes       & 5--9 & 20--60 \\
        \bottomrule
    \end{tabular}
    \caption{Configuration Parameters for Easy and Hard Difficulties in Inheritance Reasoning Topology}
    \label{tab:inherit_difficulty_presets}
    
\end{table}

\textbf{Default Reasoning:} This topology modeled a main property chain. Multiple entities could be connected to a root property. This root property implied subsequent properties. To evaluate robustness against noise, irrelevant properties (distractor nodes) were attached directly to random entities. Defeasible edges were injected onto intermediate property nodes and assigned a depth-based serial number. This serialization ensured logical order from general to specific used later translating into conversation. The hypothesis then queried a terminal property leaf.

In Table \ref{tab:parameters}, the parameters tuned to set the difficulty of default reasoning topology are listed: easy (fewer properties, objects, and hypotheses) and hard.
\begin{table}[h]
    \centering
    \small

    \begin{tabular}{lcc}
        \toprule
        \textbf{Parameter} & \textbf{Easy} & \textbf{Hard} \\
        \midrule
        Chain length            & 5--10 nodes  & 10--20 nodes \\
        Num objects             & 2--3         & 4--6         \\
        Irrelevant attributes   & 1--2         & 2--4         \\
        Sparsity factor         & 0.30--0.50   & 0.40--0.70   \\
        Max hypothesis sections & 2            & 3            \\
        \bottomrule
    \end{tabular}
    \caption{Comparison of Easy and Hard Parameters in Default Reasoning Topology}
    \label{tab:parameters}
\end{table}

\subsection{Entity Construction}
\label{sec:entity_instantiation}

The abstract graph nodes and edges from the graph generator module were populated with nonce entities and coherent attributes. The nonce entities are pronounceable but semantically null pseudowords (for example, ``Kitylu'', ``Atool'', and ``Deraw'' can be different types of ``Tool''). Entity and attribute preparations have been discussed in Appendix \ref{appendix:data_creation}. 

The property sampling was \textit{domain-aware}. A root concept dictated the pool of valid attributes (e.g., a ``Tool'' domain permits attributes like \textit{material} and \textit{utility}, whereas an ``Animal'' domain would permit \textit{habitat}). 

Additionally, for default reasoning, we sampled attributes that were distant from the attributes in the reasoning chain. These attributes were used to construct irrelevant updates. Although they described the same entity, they did not participate in any inference path leading to the hypothesis. Therefore, they should not affect its entailment.

\textbf{Template Structures: }
To generate grammatically sound sentences, properties were represented using a structured template. Each attribute category contained a pool of candidate values along with two linguistic templates (singular and plural) to support different relational contexts.

The abstract template structure was defined as follows:
\begin{verbatim}
Attribute Category: <String>
Examples: [<Value_1>, 
         <Value_2>, 
         ..., <Value_n>]
Template: "<Plural Verb Phrase> [value]"
Object Template: "<Singular Verb Phrase> 
                  [value]"
\end{verbatim}

\textbf{An Example: }
To illustrate the property instantiation, consider a graph requiring an entity assignment and a property implication within the ``Tool'' domain. 

At first, the ``material'' attribute was sampled for the ``Tool'' domain.
\begin{verbatim}
    Example Value Selected: Titanium
    Template: are made of [value]
    Object Template: is made of [value]
\end{verbatim}

Then, the placeholders in Template and Object Template were replaced with the Example Value, generating Filled Template and Filled Object Template respectively. Finally, the exact predicate structure was passed to the graph populator.  

\begin{verbatim}
Attribute Name: material
Selected Value: Titanium
Filled Template: are made of Titanium
Filled Object Predicate: is made of Titanium
\end{verbatim}

\subsection{Path Resolver Algorithm}
\label{appendix:path_resolver}

To determine the ground-truth entailment of each hypothesis, the framework employed a path-based resolver adapted from \cite{HORTY1990311}. In this formulation, conclusions are determined by which positive or negative inheritance paths are permitted by the network. When conflicting paths support opposing conclusions, the resolver applies the principle of \textit{preemption}: information inherited through a more specific class can preempt a conflicting path based on more general information. If opposing compound paths remain unpreempted, the skeptical interpretation withholds the corresponding conclusion rather than arbitrarily selecting one.

Before resolution, the logical topology was abstracted into a signed Directed Acyclic Graph (DAG), where edges denoted either positive support (supporting links) or negative evidence (defeating links). The resolver excluded hypothesis and irrelevant-property edges from logical propagation.

\begin{itemize}
    \item Supporting Links(+): ``inheritance'', ``is\_a'', ``implies'', ``has\_property'', and ``defeasible\_has\_property'' edges.
    \item Defeating Links(-): ``defeasible\_lacks\_property'' edges.
    \item Ignored Links(o): ``hypothesis'' and ``irrelevant\_property'' edges.
\end{itemize}

The evaluation then proceeded through the following core phases:

\begin{itemize}
    \item \textbf{Trimming:} The graph was first pruned to isolate the relevant nodes and edges to the specific query. The algorithm first performed a forward search from the source entity along strictly positive paths, followed by a backward search from the target property. Only nodes residing in the intersection of these valid paths were retained for evaluation.
    \item \textbf{Propagation:} The trimmed network was then evaluated in a strict topological order. This ensured that every intermediate node was fully resolved before its logical state was propagated forward. The algorithm maintained states of proven and defeated properties, carrying the deductive logic upward from the source entity.
    \item \textbf{Direct Preemption:} Any evidence directly attached to the root subject entity was treated as absolute, immediately overriding any longer, inherited reasoning chains.
    \item \textbf{Specificity Based Conflict Resolution:} When an intermediate node received conflicting logics (e.g., inheriting both a property and its negation from different parent nodes), the algorithm inspected the hierarchical relationship between the conflicting sources. If a positive path existed from source A to source B, A was identified as a more specific subclass of B. The logic originating from a more specific subclass strictly preempted the signal from a superclass.
    \item \textbf{Skeptical Inference:} If conflicting deductive paths were completely symmetric, meaning neither source node was a subclass of the other, the algorithm classified the property's state as unknown.
    \item \textbf{Default Chain Resolution:} For graphs representing linear property chains, the algorithm applied a "defaults carry through" rule. Entailment flowed through the chain unless explicitly blocked by negative defeasible evidence. Furthermore, the framework supported synthetic bypass rules, allowing the logic to skip intermediate nodes if sufficient, unblocked entry points were established earlier in the chain.
\end{itemize}

The execution of the formal PMPM strategy from \cite{HORTY1990311} is detailed throughout Algorithm \ref{alg:skeptical-inference}.

\begin{algorithm*}[h]
\caption{Skeptical Inference with Graph Trimming and Degree Ordering}
\label{alg:skeptical-inference}
\begin{algorithmic}[1]

\Function{TrimForQuery}{$\Gamma, x, y$}
    \State $ForwardSet \gets$ nodes reachable from $x$ via strictly positive links
    \State $BackwardSet \gets$ nodes from which $y$ can be reached
    \State $\Gamma' \gets ForwardSet \cap BackwardSet$
    \State \Return $\Gamma'$
\EndFunction

\Statex

\Function{SelectNextDegree}{$\Gamma', P$}
    \State $N \gets \emptyset$
    \ForAll{node $v \in \Gamma'$ such that $v \notin P$}
        \If{all predecessors of $v$ are contained in $P$}
            \State $N \gets N \cup \{v\}$
        \EndIf
    \EndFor
    \State \Return $N$
\EndFunction

\Statex

\Function{Query}{$\Gamma, x, y$}
    \State $\Gamma' \gets \Call{TrimForQuery}{\Gamma, x, y}$
    \State $P \gets \emptyset$
    \State $State(x) \gets \text{True}$

    \While{$P \neq \Gamma'$}
        \State $N \gets \Call{SelectNextDegree}{\Gamma', P}$

        \ForAll{node $w \in N$}

            \Comment{1. Direct evidence overrides inherited evidence}
            \If{direct positive link $(x \to w)$ exists in $\Gamma$}
                \State $State(w) \gets \text{True}$

            \ElsIf{direct negative link $(x \not\to w)$ exists in $\Gamma$}
                \State $State(w) \gets \text{False}$

            \Else
                \Comment{2. Collect inherited evidence from valid parents}

                \State $PosParents \gets
                \{p \mid State(p)=\text{True}
                \land (p \to w)\text{ exists}\}$

                \State $NegParents \gets
                \{p \mid State(p)=\text{True}
                \land (p \not\to w)\text{ exists}\}$

                \If{$PosParents \neq \emptyset$
                    \textbf{ and } $NegParents = \emptyset$}

                    \State $State(w) \gets \text{True}$

                \ElsIf{$NegParents \neq \emptyset$
                    \textbf{ and } $PosParents = \emptyset$}

                    \State $State(w) \gets \text{False}$

                \ElsIf{$PosParents \neq \emptyset$
                    \textbf{ and } $NegParents \neq \emptyset$}

                    \Comment{3. More specific information defeats more general information}
                    \State $State(w) \gets
                    \Call{ResolveSpecificity}{PosParents, NegParents}$

                \Else
                    \State $State(w) \gets \text{Unknown}$
                \EndIf
            \EndIf

        \EndFor

        \State $P \gets P \cup N$
    \EndWhile

    \State \Return $State(y)$
\EndFunction

\end{algorithmic}
\end{algorithm*}

The resolver returns one of three states for the target property node: entailed (True), defeated (False), or undetermined (Unknown).

\subsection{Natural Language Generation}

To translate the populated graph into readable text, a natural language generation module mapped each directed edge to a coherent English sentence. This was achieved using predefined structural templates that defined the grammatical interaction between a source node and a target node based on the edge type. 

Common sentence templates include: 

\begin{table}[h]
\centering
\caption{Natural language templates used for different edge types.}
\label{tab:edge_templates}
\resizebox{\columnwidth}{!}{%
\begin{tabular}{p{0.30\columnwidth} p{0.62\columnwidth}}
\hline
\textbf{Edge Type} & \textbf{Template} \\
\hline

\texttt{is\_a}
& \texttt{\{subject\} is a \{object\}.} \\
\hline
\texttt{implies}
& Objects that \texttt{\{subject\_predicate\}} generally 
  \texttt{\{object\_predicate\}}. \\
\hline
\texttt{conflict}
& However, \texttt{\{subject\}} does not \texttt{\{object\}}. \\
\hline
\texttt{has\_attribute}
& \texttt{\{subject\}} also \texttt{\{object\_predicate\}}. \\
\hline
\texttt{inheritance}
& \texttt{\{subject\}} is a kind of \texttt{\{object\}}. \\
\hline
\texttt{defeasible has property}
& \texttt{\{subject\}} \texttt{\{object\_predicate\}}. \\
\hline
\texttt{defeasible lacks property}
& \texttt{\{subject\}} is not \texttt{\{object\}}. \\
\hline
\texttt{hypothesis edge}
& Does \texttt{\{subject\}} \texttt{\{object\_predicate\}}? \\

\hline
\end{tabular}%
}
\end{table}

During sentence generation, the module injected the instantiated entity labels and filled property predicates into these sentence templates. For example, given a nonce entity (\textit{Zibber}), a base concept (\textit{Tool}), and an instantiated property (\textit{made of Titanium}), the graph edges are translated as follows: 

\begin{itemize}
    \item {\textbf{Inheritance Edge Translation:}}\\
    \textit{Mapping:} \{subject: Zibber\} is a \{object: Tool\}. \\
    \textit{Final Output:} ``The Zibber is a Tool.''
    
    \item {\textbf{Property Assignment Edge Translation:}}\\
    \textit{Mapping:} \{subject: Zibber\} \{object\_predicate: is made of Titanium\}. \\
    \textit{Final Output:} ``The Zibber is made of Titanium.''
    
    \item {\textbf{Implication Edge Translation:}}\\
    \textit{Mapping:} Objects that \{subject\_predicate: are Tools\} generally \{object\_predicate\_plural: are made of Titanium\}. \\
    \textit{Final Output:} ``Objects that are Tools generally are made of Titanium.''
\end{itemize}


\subsection{Conversation Simulation}


After the natural language realization of every graph edge, the conversation simulator determines the order in which edges of the reasoning structure are disclosed. After each disclosure, the resolver reevaluates the hypothesis. 

An initial premise was first constructed from the deterministic backbone of the graph: consisting of the ``inheritance'' and ``has-attribute'' edges. This premise, along with the hypothesis, was presented to the resolver before any defeasible evidence was introduced. It established the base structure against which all subsequent updates would be evaluated.

Defeasible edges were then revealed according to a controlled schedule. Each defeasible edge was assigned a serial index according to its position in the reasoning structure, so that lower indices corresponded to more general information and higher indices to more specific information. Two reveal strategies were used for defeasible edges: \textit{serial reveal} and \textit{randomized reveal}. 

\begin{itemize}
    \item \textit{Serial Reveal:} In this configuration, edges were disclosed in ascending order of their assigned serial indices. That means the revealed information became more and more specific. 

    \item \textit{Randomized Reveal:}  An alternative randomized reveal mode was also supported to mitigate the risk of models exploiting positional patterns rather than inferential reasoning. In that mode, defeasible edges for a given subject were shuffled into a random reveal order. The resolver was supplied only with the edges explicitly revealed up to that point.
\end{itemize}


After each defeasible edge was introduced, the hypothesis was re-evaluated using all evidence revealed up to that turn. For inheritance reasoning, this produced only the updated entailment label: \textit{yes}, \textit{no}, or \textit{unknown}. The effect of an update was determined by comparing the hypothesis state before and after the reveal. An update was strengthening if it increased support for the hypothesis, weakening if it reduced support, and no effect if the resolved hypothesis state remained unchanged. 

Moreover, a positive or negative defeasible edge did not necessarily imply strengthening or weakening by itself. For example, a negative defeasible edge introduced on a path whose inheritance had already been solved would leave the hypothesis unchanged (no effect). 

Inheritance conversations used only the entailment label, while the explicit strengthening, weakening, and no-effect labels were used for the belief-update questions in default reasoning.


\section{Default Reasoning}
\label{appendix:default_reasoning}



In this section, we have discussed which of the benchmark problems from \cite{lifschitz1988benchmark} have been addressed by every question type in Default reasoning.

\textbf{Type 1: Direct Chain Manipulation}
\label{subsec:type1}

Consider an initial premise: 

``Heavy blocks are normally located on the table.'' 

\begin{itemize}
    \item \textit{In Basic Default Reasoning} (Benchmark A1), introducing the fact:

    ``Block A is heavy'' 

    leads to the conclusion:

    ``Block A is on the table.''

    \item \textit{Default Reasoning with Several Defaults} (Benchmark A3) extends it for several defaults. 

    \item \textit{Default Reasoning with a Disabled Default} (Benchmark A4) tests the revocation of this conclusion by introducing an overriding fact: 
    
    ``Block A is an exception to this rule.''
\end{itemize}

The NL Generator traced the primary logical chain. Introducing new facts either reinforced the current hypothesis or broke the chain. When generating a negative update, NL Generator randomly alternated between a \textit{direct negation} (``The object does not have the property'') and \textit{exception phrasing} (``The object is an exception to the rule''). If the parent category has already been established as negative earlier in the chat, the system forcefully overrode the exception phrasing to prevent contradiction. 

\textbf{Type 2: Irrelevant Information}
\label{subsec:type2}

To understand \textit{Default Reasoning with Irrelevant Information} (Benchmark A2), consider the generator has concluded: 

``Block A is on the table''. 

Introducing a new, unrelated fact such as 

``Block A is red'',

should not affect the prior conclusion. 

The NL Generator generated two distinct variants of irrelevance to probe the model's robustness. ``Safe'' irrelevance introduced completely benign and unrelated traits. ``Trap'' irrelevance dynamically scanned the active logic graph and synthesized distractor properties that share overlapping vocabulary or negative prefixes with valid chain nodes. In all Type 2 updates, the programmed ground-truth registered ``no effect'' on the underlying hypothesis.




    
    
    









\textbf{Type 3: Source-Priority Conflicts}
\label{subsec:type4}

\begin{itemize}
    \item \textit{Priorities between Defaults} (Benchmark A9) involves conflicting sources(Jack and Mary). If Jack asserts, 
    
    ``Block A is not on the table,'' 
    
    while Mary asserts, 
    
    ``Block A is on the table.'' 
    
    Because the environment dictates that Mary is a more reliable source than Jack, her claim takes precedence, and the final conclusion must side with her.

\end{itemize}

NL Generator evaluated complex belief revision by orchestrating disputes between simulated agents (``Expert A'' and ``Expert B''). A known priority ordering was announced in the pretext once: 

    "Expert A is generally more reliable than Expert B."
    
    The implementation had several mechanisms:

\begin{itemize}
    \item \textbf{Dual Conflict Topologies:} Realization of two distinct shapes of disagreement:
    
    The first is a \textit{direct contradiction}, where the high-priority and low-priority agents argued over the exact same node. 
    
    The second is a \textit{structural bypass}. In a bypass, the low-priority agent claimed that a critical connective rule is blocked. 
    
    Rather than arguing the point, the high-priority agent introduced a ``shortcut'' rule that logically routed around the blocked node (ensuring the bypass source index was strictly before the block, and the target index was strictly after), successfully restoring the hypothesis path.

    \begin{figure}[h]
    \centering
    \includegraphics[width=0.9\linewidth]{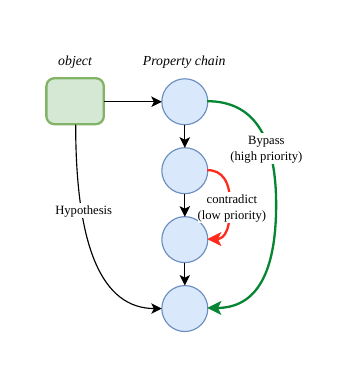}
    \caption{Conflict: Contradict and Bypass}
    \label{fig:tree_preset}
    \end{figure}
    \item \textbf{Temporal Ordering:} Randomizing the order in which the agents speak:
    
    If the low-priority agent spoke first, the resolver registered a drop in confidence, followed by a nonmonotonic recovery when the high-priority agent corrected the record. 
    
    Conversely, if the high-priority agent spoke first, the system acted as an epistemic shield. When the low-priority agent disagreed, the resolver computed ``no effect''.
    
    \item \textbf{Redundancy Filtering:} If a lower-priority source subsequently echoed a confirmation already established by a higher-priority source, the algorithm enforced a strict ``no effect''. 
    
    \item \textbf{Subject Isolation:} Entities subjected to Type 3 source-priority conflicts were disjoint from entities undergoing Type 1 or 2 scenarios to ensure the model's success or failure is a function of priority resolution.
\end{itemize}

In Table \ref{tab:benchmark_mapping}, we have briefly discussed the mapping.

\begin{table}[h]
    \centering
    \small
    \caption{Mapping of Question Types to \cite{lifschitz1988benchmark}'s Benchmark Equivalents}
    \label{tab:benchmark_mapping}
    \renewcommand{\arraystretch}{1.4} 
    \begin{tabular}{@{}p{0.45\linewidth} p{0.5\linewidth}@{}}
        \toprule
        \textbf{Question Type} & \textbf{Benchmark Equivalents} \\
        \midrule
        \textbf{Type 1:} Direct Chain Manipulation & \textbf{A1} (Basic Default Reasoning) \newline 
        \textbf{A3} (Default Reasoning with Several Defaults) \newline
        \textbf{A4} (Disabled Default) \\
        
        \textbf{Type 2:} Irrelevant Information & \textbf{A2} (Default Reasoning with Irrelevant Information) \\
        
        
        \textbf{Type 3:} Source-Priority Conflicts & \textbf{A9, A10} (Priorities between Defaults) \\
        \bottomrule
    \end{tabular}
\end{table}

\section{Inheritance Reasoning} 
\label{appendix:inheritance_reasoning}


In Inheritance reasoning, a target property was implied by the root entity/domain and was inherited by its descendants. A negative defeasible update (defeasible\_lack\_property edge) could block a property inherited from a more general entity. Whereas, a positive update to a more specific entity could again restore the property.

Figure \ref{fig:tree_preset} shows the real-life hierarchy mimics we have used as presets.

\begin{figure*}[h]
    \centering
    \includegraphics[width=0.9\linewidth]{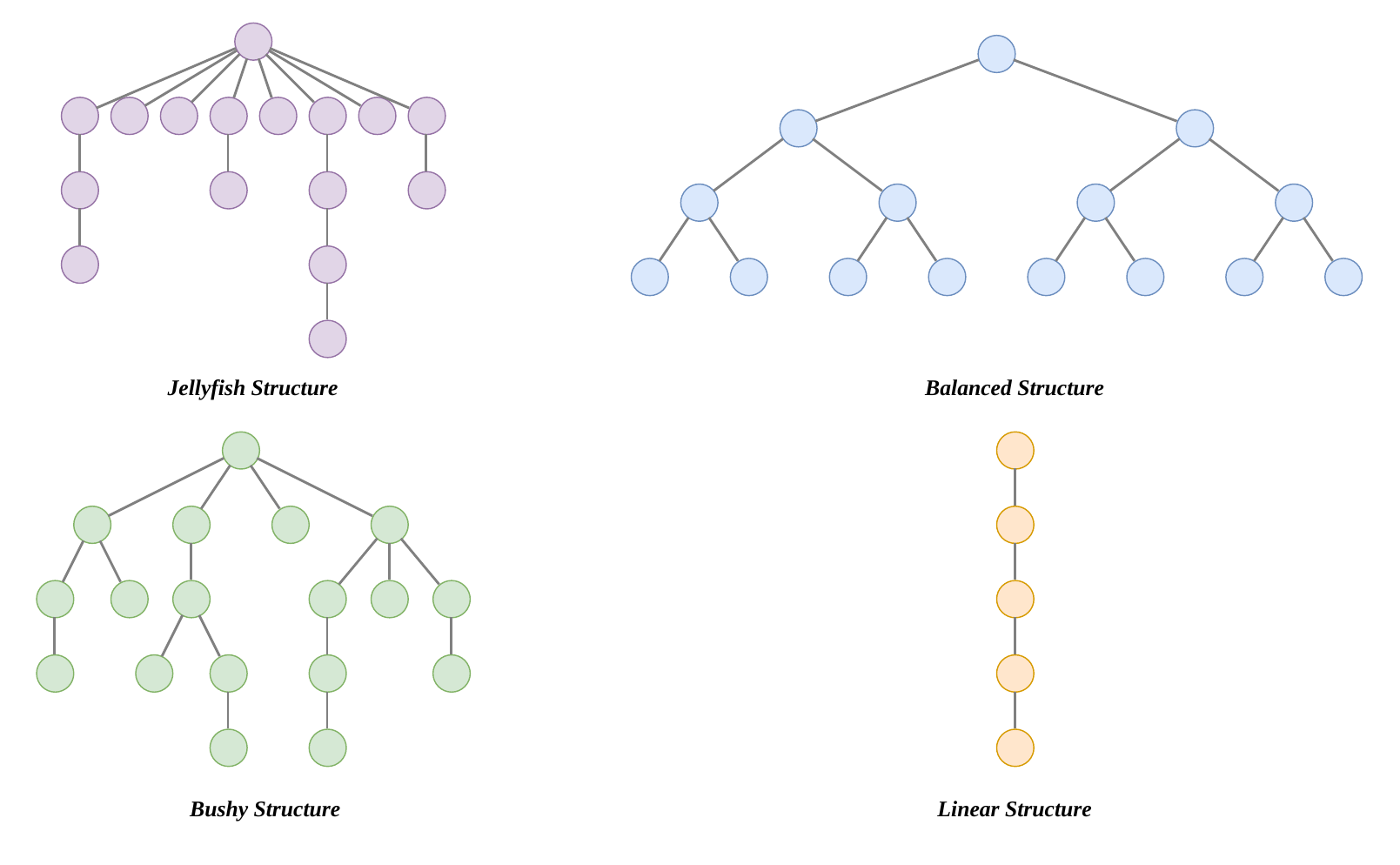}
    \caption{Inheritance: Real Life Hierarchy Mimic}
    \label{fig:tree_preset}
\end{figure*}

\section{Data Creation Process}
\label{appendix:data_creation}
This section contains the overall architecture of the whole data creation process.
\subsection{Nonce Word Generation} 
\label{appendix:data_creation_nonce}

There was the possibility that if we used real and sensical words as entity, models could answer using memorized real-world knowledge instead of reasoning over the premises provided in the conversation. Because nonce entities carry no predefined semantic associations, the resulting examples directly tested the defeasible reasoning process.

To populate the entity nodes of the reasoning structure, we generated the nonce words using character-level Markov n-gram models trained on different seed corpora following the UniPseudo paradigm of \cite{new2024unipseudo}. We followed the CGCA approach of \cite{konig2020using} to avoid bleed between word-initial and word-final contexts. In this approach, the model's transitions were bucketed by relative position (initial, medial, final). 

Afterwards, transition probabilities were smoothed using interpolated Kneser–Ney following \cite{chen1999empirical}, which prevents zero-probability assignments on unseen continuations and supports candidate scoring. Each candidate was additionally subject to a legality check \cite{konig2020using}.

To train the model, we prepared different reference corpora: reference nouns and adjectives, synthetic phonotactic syllables, domains with Datamuse expansion, etc. We also sampled nonce words, locking the suffix for noun and adjective-shaped nonce words. Finally, we filtered out real English words, repeated characters, and unpronounceable consonant strings.

\subsection{Property Fetching and Enrichment}

At first, we prepared a list of domain names. To fetch properties given the domain names, we queried the normalized ConceptNet DB\footnote{\url{https://huggingface.co/datasets/cstr/conceptnet-normalized-multi/tree/main}}, WikiData\footnote{\url{https://www.wikidata.org/wiki/Wikidata:REST_API}}, and Datamuse API\footnote{\url{https://www.datamuse.com/api/}}. Then we categorized the properties into groups. For instance, if a domain animal'' has two properties named fly'' and swim'', we assigned them to the group of capability''. Moreover, we used a Large Language Model to augment more properties under each group, augment new property groups, and enrich the domain-attribute mapping.

We represented each such group as ``attribute'' and properties under a group as ``values''. To use these properties in a sentence, we need to prepare templates. For instance, ``fly'' is a value under the attribute ``capability''. Now, the sentence should be ``[Subject] can fly''. Here, the object predicate was stored as a template (``can [value]''). Values can be different capabilities in this case. Table \ref{tab:domain_attributes} lists attribute counts and sample attributes for each domain.  

\begin{table}[h]
\centering
\begin{tabular}{|p{0.25\columnwidth}|p{0.15\columnwidth}|p{0.43\columnwidth}|}
\hline
\textbf{Domain Name} & \textbf{Attr. Count} & \textbf{Sample Attributes} \\ \hline

Animal & 32 & Sound, Behavior, Size, Weight \\ \hline
Plant & 30 & Location, Origin, Age \\ \hline
Tool & 25 & Material, Utility, Affordance \\ \hline
Food & 24 & Texture, Flavor, Appearance \\ \hline
Vehicle & 25 & Sound, Origin, State \\ \hline
Compound & 21 & Color, Shape, Size \\ \hline
Object & 24 & State, Utility, Role \\ \hline
Material & 22 & Composition, Temperature, Age \\ \hline
Micro organism & 18 & Origin, Size, State\\ \hline
Infrastructure & 24 & Location, Shape, Weight \\ \hline

\end{tabular}
\caption{Domain Names and their corresponding Attribute Counts \textit{(Attr. Count)}.}
\label{tab:domain_attributes}
\end{table}

\begin{table*}[t]
\centering
\small
\setlength{\tabcolsep}{5pt}
\begin{tabular}{llcccccc}
\toprule
\textbf{Model} & \textbf{Config} &
\textbf{Mean} & \textbf{Std} & \textbf{CV\,(\%)} &
\textbf{H} & \textbf{$p$} & \textbf{Sig.} \\
\midrule
\multicolumn{8}{l}{\textit{Default Reasoning, Easy}} \\
\quad Qwen3-32B
  & Default & 0.733 & 0.025 & 3.36 & 16.41 & 0.630 & No \\
\quad Gemma-4-12B
  & Default  & 0.791 & 0.027 & 3.47 & 20.63 & 0.358 & No \\
\quad Gemma-4-26B-A4B
  & Default  & 0.790 & 0.026 & 3.24 & 16.72 & 0.609 & No \\
\quad Gemma-4-31B
  & Default  & 0.842 & 0.026 & 3.10 & 28.75 & 0.070 & No \\
\quad Gemma-4-E4B
  & Default  & 0.467 & 0.028 & 5.96 & 12.46 & 0.865 & No \\
\quad gpt-5.1
  & Default  & 0.702 & 0.026 & 3.73 & 16.59 & 0.617 & No \\
\quad Llama-3.1-70B
  & Default  & 0.659 & 0.024 & 3.57 & 17.64 & 0.546 & No \\
\quad Llama-3.1-8B
  & Default  & 0.528 & 0.036 & 6.90 & 21.19 & 0.327 & No \\
\midrule
\multicolumn{8}{l}{\textit{Linear Inheritance, Easy}} \\
\quad Qwen3-32B
  & Linear  & 0.893 & 0.036 & 4.01 & 22.99 & 0.238 & No \\
\quad Gemma-4-12B
  & Linear  & 0.897 & 0.029 & 3.25 & 17.95 & 0.526 & No \\
\quad Gemma-4-26B-A4B
  & Linear  & 0.867 & 0.034 & 3.93 & 13.95 & 0.787 & No \\
\quad Gemma-4-31B
  & Linear  & 0.932 & 0.032 & 3.44 & 19.29 & 0.439 & No \\
\quad Gemma-4-E4B
  & Linear  & 0.683 & 0.054 & 7.86 & 17.26 & 0.572 & No \\
\quad gpt-5.1
  & Linear  & 0.921 & 0.033 & 3.62 & 18.74 & 0.474 & No \\
\quad Llama-3.1-70B
  & Linear  & 0.908 & 0.029 & 3.20 & 14.98 & 0.724 & No \\
\quad Llama-3.1-8B
  & Linear  & 0.665 & 0.046 & 6.88 & 21.16 & 0.328 & No \\
\bottomrule
\end{tabular}
\caption{Per-cell Kruskal--Wallis results across 20 seeds.
Mean and Std are computed over per-seed accuracy scores.
CV = coefficient of variation (Std / Mean $\times$ 100).
H = Kruskal--Wallis statistic; $p$ = asymptotic $p$-value
under $\chi^2$ approximation with 19 degrees of freedom.
No cell is significant at $\alpha = 0.05$.}
\label{tab:kruskal_wallis}
\end{table*}

\section{Data Examples}
\label{appendix:data_samples}

\subsection{Default Reasoning}

\begin{figure*}[h]
    \centering
    \includegraphics[width=0.9\linewidth]{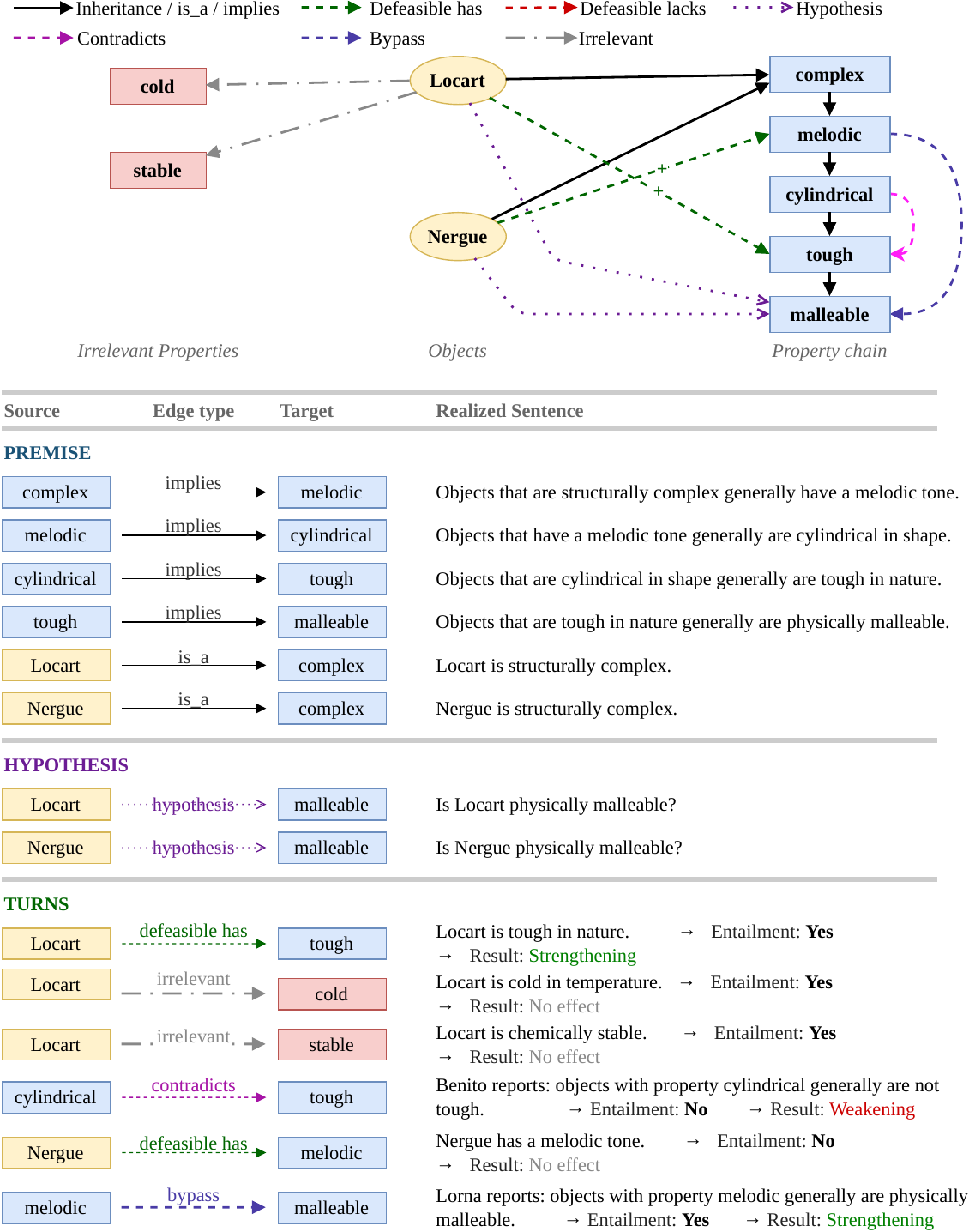}
    \caption{Default Reasoning Example}
    \label{fig:default_example}
\end{figure*}

In Figure \ref{fig:default_example}, an example graph on default reasoning is shown.

\subsection{Linear Inheritance}

\begin{figure*}[h]
    \centering
    \includegraphics[width=0.9\linewidth]{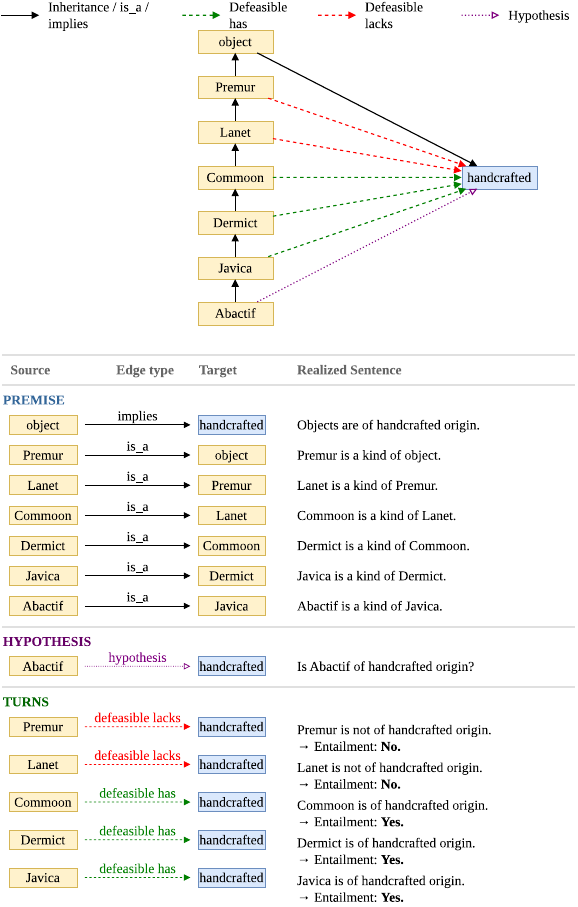}
    \caption{Linear Inheritance Example}
    \label{fig:linear_example}
\end{figure*}

In Figure \ref{fig:linear_example}, an example graph on linear inheritance is shown.

\subsection{Tree Inheritance}

\begin{figure*}[h]
    \centering
    \includegraphics[width=0.8\linewidth]{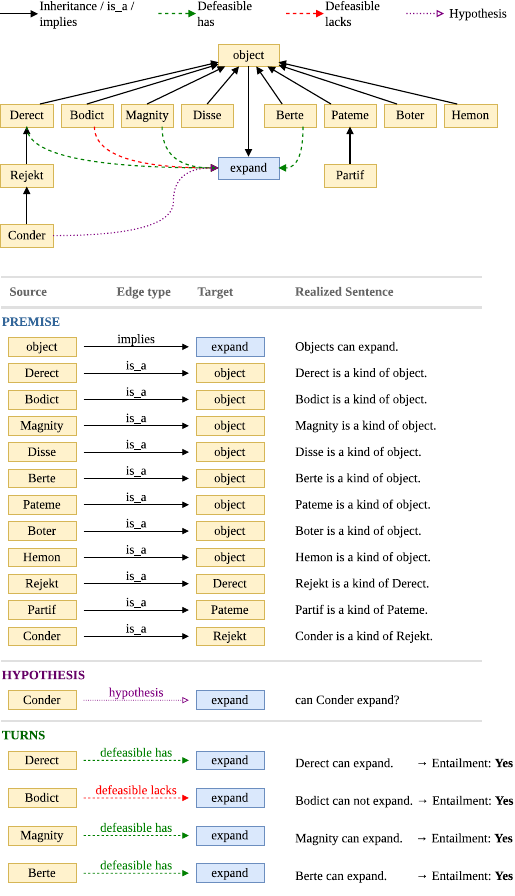}
    \caption{Tree Inheritance Example}
    \label{fig:tree_example}
\end{figure*}

In Figure \ref{fig:tree_example}, an example graph on tree inheritance is shown.


\label{appendix:stat_sig_models}


\section{Seeded Sample Variance: Statistical Details}
\label{appendix:stat_sig_reseed}

This section provides full statistical details for the seeded
sample variance analysis. We report the Intraclass Correlation Coefficient (ICC) pooled
across all configurations and per-configuration Kruskal--Wallis
tests of seed-level accuracy distributions.

\subsection*{Experimental Setup}

We run 20 independent random seeds per configuration,
generating 30 conversations per seed.
Eight models are evaluated: \texttt{gpt-5.1}, \texttt{Qwen3-32B}, 
Gemma-4 family (\texttt{Gemma-4-E4B}, \texttt{Gemma-4-12B}, \texttt{Gemma-4-26B-A4B MoE}, \texttt{Gemma-4-31B}), \texttt{Llama-3.1-70B-Instruct} and \texttt{Llama-3.1-8B-Instruct}, representing frontier closed, open large-scale to small model tiers, and a family of newer architecture models. Two configurations are tested: \textit{Default Reasoning}
and \textit{Linear Inheritance} against the easy tier difficulty.
Accuracy is computed at the turn level against formally verified
ground truth labels for each seed independently.

\subsection*{Intraclass Correlation Coefficient}

To quantify the consistency of accuracy scores across seeds, we compute the two-way random-effects ICC(2,1) treating seeds as the replicate factor and configurations as the grouping factor.
The ICC is estimated on a $16 \times 20$ matrix (16 cells
= 8 models $\times$ 2 configurations; 20 seeds), with
bootstrap confidence intervals computed over 1,000 resamples.

As shown in Table \ref{tab:icc}, the pooled ICC of $0.949$ (95\,\% CI $[0.9382,\,0.9631]$) indicates \emph{excellent} consistency across seeds under all tested configurations~\citep{koo2016guideline}. The upper bound of the CI does not fall below $0.93$, confirming that this conclusion is robust to bootstrap sampling variability.

\begin{table}[t]
\centering
\small
\begin{tabular}{lc}
\toprule
\textbf{Statistic} & \textbf{Value} \\
\midrule
ICC(2,1) & 0.949 \\
95\,\% CI lower bound & 0.9382 \\
95\,\% CI upper bound & 0.9631 \\
Number of cells & 16 \\
Number of seeds per cell & 20 \\
Bootstrap resamples & 1,000 \\
\bottomrule
\end{tabular}
\caption{Pooled ICC(2,1) across all model--configuration cells.}
\label{tab:icc}
\end{table}

\subsection*{Kruskal--Wallis Tests}

For each model--configuration cell, we additionally run a Kruskal--Wallis H-test with the null hypothesis that the 20 per-seed accuracy scores are drawn from
the same distribution.
Unlike the ICC, which measures overall consistency, the
Kruskal--Wallis test is sensitive to any systematic ordering
or drift across seeds within a cell.
A non-significant result ($p > 0.05$) provides direct evidence
that no seed produces systematically higher or lower accuracy
than others.

\noindent
All 16 cells are non-significant (minimum $p = 0.07$,
maximum $H = 28.75$) considering $ \alpha = 0.05 $, providing no evidence that any seed produces systematically different results within a given
configuration. The coefficient of variation remains below 8\,\% in all cells, with \texttt{Gemma-4-31B} showing the lowest
variance (CV $\leq 3.10$\,\%) and \texttt{Gemma-4-E4B} on
linear inheritance showing the highest (CV $= 7.86$\,\%),
still well within the range expected from sampling variability
alone at $n = 30$ conversations per seed.

Taken together, the ICC and Kruskal--Wallis results confirm
that \textsc{DeReLab} evaluations are statistically
reproducible under reseeding: different seeds generate
statistically equivalent datasets from the same configuration,
and no seed produces a systematically advantaged or
disadvantaged evaluation.

\begin{table*}[t]
\centering
\small
\setlength{\tabcolsep}{5pt}
\begin{tabular}{llrrrrrr}
\toprule
\multirow{2}{*}{\textbf{Model}} &
\multirow{2}{*}{\textbf{Diff}} &
\multirow{2}{*}{\textbf{n}} &
\multicolumn{2}{c}{\textbf{Distance effect}} &
\multicolumn{2}{c}{\textbf{Per-hop } $\hat\beta$} &
\textbf{Calibration} \\
\cmidrule(lr){4-5}\cmidrule(lr){6-7}
& & & $\rho$ & $p$ & $\hat\beta$ & $p$ & $\Delta$Brier \\
\midrule
Llama-3.1-8B    & Easy & 150 & $-0.230$ & 0.005      & $-0.069$ & ${<}0.001$ & $+0.112$ \\
Llama-3.1-8B    & Hard & 294 & $-0.255$ & ${<}0.001$ & $-0.069$ & ${<}0.001$ & $+0.131$ \\
\midrule
Llama-3.1-70B   & Easy & 150 & $-0.153$ & 0.062      & $-0.056$ & ${<}0.001$ & $+0.080$ \\
Llama-3.1-70B   & Hard & 294 & $-0.207$ & ${<}0.001$ & $-0.056$ & ${<}0.001$ & $+0.100$ \\
\midrule
Gemma-4-E4B     & Easy & 150 & $-0.037$ & 0.651      & $-0.021$ & 0.114      & $-0.078$ \\
Gemma-4-E4B     & Hard & 294 & $+0.072$ & 0.217      & $-0.021$ & 0.114      & $-0.067$ \\
\midrule
Gemma-4-12B     & Easy & 150 & $-0.140$ & 0.087      & $-0.064$ & ${<}0.001$ & $-0.032$ \\
Gemma-4-12B     & Hard & 294 & $-0.156$ & 0.007      & $-0.064$ & ${<}0.001$ & $+0.146$ \\
\midrule
Gemma-4-31B     & Easy & 150 & $-0.273$ & ${<}0.001$ & $-0.061$ & ${<}0.001$ & $-0.028$ \\
Gemma-4-31B     & Hard & 294 & $-0.298$ & ${<}0.001$ & $-0.061$ & ${<}0.001$ & $+0.157$ \\
\midrule
Gemma-4-MoE     & Easy & 150 & $-0.203$ & 0.013      & $-0.039$ & 0.014      & $+0.131$ \\
Gemma-4-MoE     & Hard & 294 & $-0.184$ & 0.002      & $-0.039$ & 0.014      & $+0.128$ \\
\midrule
GPT-5.1         & Easy & 150 & $+0.015$ & 0.860      & $+0.029$ & 0.138      & $-0.031$ \\
GPT-5.1         & Hard & 294 & $+0.130$ & 0.026      & $+0.029$ & 0.138      & $-0.170$ \\
\midrule
Qwen3-32B       & Easy & 150 & $-0.105$ & 0.203      & $-0.026$ & 0.095      & $+0.068$ \\
Qwen3-32B       & Hard & 294 & $-0.030$ & 0.607      & $-0.026$ & 0.095      & $-0.005$ \\
\midrule
Qwen3-32B*      & Easy & 150 & $-0.042$ & 0.609      & $-0.017$ & 0.207      & $-0.003$ \\
Qwen3-32B*      & Hard & 151 & $-0.010$ & 0.903      & $-0.017$ & 0.207      & $+0.049$ \\
\bottomrule
\end{tabular}
\caption{On-path confidence analysis. $\rho$ is the Spearman correlation
between update distance and gold-answer confidence (negative = confidence
erodes with distance). $\hat\beta$ is the mixed-effects coefficient giving
expected $\Delta\mathrm{gold\_conf}$ per additional hop, controlling for
conversation-level baseline. $\Delta$Brier is the change in Brier score from
short ($d{\leq}2$) to long ($d{>}2$) chains; positive values indicate
worsening calibration at greater depth.
Qwen3-32B* uses extended thinking mode.}
\label{tab:path_onpath}
\end{table*}

\section{Path Confidence Analysis: Full Statistical Details}
\label{appendix:path_confidence}

This appendix provides complete statistical results for the
inference-path confidence analysis summarized in
\S\ref{sec:path_confidence} of the main paper. This analysis is restricted to the tree inheritance paradigm where we want to check if there is a correlation between the graph structural distance of an incoming update and the hypothesis node and its effect on the model confidence.
 
\subsection*{Metrics and Statistical Tests}

\paragraph{Structural distance $d$.}
Each \texttt{update\_answer} turn is annotated with $d$, the shortest-path
distance in the tree between the update node and the hypothesis node.
Distance $d{=}1$ means the update directly modifies a property the hypothesis entails; larger values require more inferential hops to reach the hypothesis.

\paragraph{On-path vs.\ off-path turns.}
A turn is \emph{on-path} if the update's logical effect is \textit{strengthen} or \textit{weaken} — it is causally relevant to the hypothesis. It is \emph{off-path} if the effect is \textit{no effect} — the update belongs to a sibling branch whose changes do not propagate to the hypothesis node. These conditions are non-overlapping and form two separate analyses.

\paragraph{Gold confidence (\textbf{gold\_conf}).}
Each model response includes a constrained probability distribution
$\mathbf{p} \in \Delta^{|\mathcal{Y}|}$ over the answer label set~$\mathcal{Y}$.
We define
\[
  \mathrm{gold\_conf} \;=\; p_{y^{*}},
\]
where $y^{*}$ is the ground-truth label.
This measures how much probability mass the model places on the correct answer,
regardless of which label it actually chose.

\paragraph{Predicted confidence (\textbf{pred\_conf}).}
\[
  \mathrm{pred\_conf} \;=\; p_{\hat{y}},
\]
where $\hat{y} = \arg\max_y p_y$ is the model's predicted label.
This measures the model's self-assessed certainty in its own answer.

\paragraph{Confidence delta ($\Delta\mathrm{gold\_conf}$).}
Raw gold confidence conflates per-sample baseline difficulty with path-length
effects.
We normalise by subtracting the initial confidence before any updates in the
conversation:
\[
  \Delta\mathrm{gold\_conf}(t) \;=\;
  \mathrm{gold\_conf}(t) - \mathrm{gold\_conf}(t_0),
\]
where $t_0$ is the first \texttt{initial\_answer} turn in the conversation.
Negative $\Delta$ indicates confidence erosion relative to baseline.

\paragraph{Entropy and margin.}
We track two complementary uncertainty measures.
Shannon entropy $H(\mathbf{p}) = -\sum_{y} p_y \log p_y$ (in nats) captures
overall spread across the label set.
The decision margin $m = p_{(1)} - p_{(2)}$, where $p_{(k)}$ is the $k$-th
largest probability, captures the strength of the model's top preference.
High entropy and low margin both indicate uncertainty; they can diverge when the
distribution has a long tail.

\paragraph{Intrusion error.}
On an off-path turn, an \emph{intrusion error} occurs when the model changes its
answer from the previous turn \emph{and} the new answer is wrong:
\begin{equation*}
\begin{split}
  \mathrm{intrusion}(t) &= 
    \mathbf{1}[\text{effect} = \text{no effect}] \\
    &\quad \cdot \mathbf{1}[\hat{y}_t \neq \hat{y}_{t-1}] \\
    &\quad \cdot \mathbf{1}[\hat{y}_t \neq y^{*}_t]
\end{split}
\end{equation*}
The \emph{intrusion rate} is the fraction of off-path turns in a group that
satisfy this condition.
It directly measures how often a model revises its belief in response to
causally irrelevant evidence.

\paragraph{Statistical tests.}
We use three tests to measure how confidence changes with path length.
\textbf{Spearman $\rho$} captures whether confidence rises or falls
monotonically as $d$ increases.
\textbf{Mixed-effects regression} fits $\Delta\mathrm{gold\_conf}$ against
$d$ with a per-conversation random intercept, so the coefficient $\hat{\beta}$
gives the expected confidence change per additional hop while controlling for
each conversation's baseline difficulty.
\textbf{Brier scores} compare calibration quality on short ($d \leq 2$)
versus long ($d > 2$) chains — a rising score means the model's probability
estimates become less reliable as chains grow.

\subsection{On-Path Results}
\label{sec:onpath_results}

Table~\ref{tab:path_onpath} shows three measures of how reasoning quality
changes as the update moves further from the hypothesis in the inheritance
tree: the Spearman correlation $\rho$ between distance and gold-answer
confidence, the per-hop regression coefficient $\hat\beta$, and the change
in Brier score from short to long chains ($\Delta$Brier).

\paragraph{Five of nine models degrade with distance.}
Llama-3.1-8B, Llama-3.1-70B, Gemma-4-12B, Gemma-4-31B, and Gemma-4-MoE
all show consistent negative $\rho$ and significant negative $\hat\beta$
($p < 0.05$ after controlling for difficulty).
The effect is strongest for Gemma-4-31B ($\rho = -0.298$, $\hat\beta =
-0.061$ on hard) and the Llama models ($\hat\beta = -0.069$ and $-0.056$
respectively), meaning each additional inferential hop costs these models
up to 6--7 percentage points of correct-label confidence.
Gemma-4-MoE shows a similar but weaker pattern ($\hat\beta = -0.039$).

\paragraph{GPT-5.1, Qwen3, and Gemma-4-E4B are robust.}
None of these three show a significant distance effect, and their
$\Delta$Brier values are zero or negative — meaning calibration does not
degrade and, for GPT-5.1 on the hard split, actually improves at longer
chains ($\Delta\text{Brier} = -0.170$).
Gemma-4-E4B is a special case: it is distance-robust but has the worst
absolute Brier scores in the table (0.715 short-path, hard), indicating
that its probability estimates are unreliable throughout, not just at depth.

\paragraph{Hard chains amplify calibration degradation.}
For most models, the easy split shows moderate or even negative $\Delta$Brier,
while the hard split is consistently positive and substantially larger.
Gemma-4-31B, for example, has near-zero Brier on both short and long
easy chains, but its long-chain hard Brier rises to 0.201 ($\Delta = +0.157$),
suggesting that the compounding difficulty of harder graphs accelerates
the breakdown of calibrated confidence at depth.

\begin{table}[h]
\centering
\small
\setlength{\tabcolsep}{5pt}
\begin{tabular}{llrrr}
\toprule
\textbf{Model} & \textbf{Diff} &
\textbf{Intrus.} &
\textbf{SR} &
\textbf{Off-path acc.} \\
\midrule
Llama-3.1-8B    & Easy & 36.2\,\% & $0.68\times$ & 47.8\,\% \\
Llama-3.1-8B    & Hard & 31.9\,\% & $0.82\times$ & 36.3\,\% \\
\midrule
Llama-3.1-70B   & Easy & 15.2\,\% & $0.80\times$ & 63.0\,\% \\
Llama-3.1-70B   & Hard &  8.1\,\% & $0.83\times$ & 32.0\,\% \\
\midrule
Gemma-4-E4B     & Easy &  4.2\,\% & $0.98\times$ & 45.1\,\% \\
Gemma-4-E4B     & Hard &  5.6\,\% & $1.00\times$ & 31.1\,\% \\
\midrule
Gemma-4-12B     & Easy &  1.1\,\% & $1.03\times$ & 94.7\,\% \\
Gemma-4-12B     & Hard &  1.4\,\% & $0.99\times$ & 80.7\,\% \\
\midrule
Gemma-4-31B     & Easy &  0.3\,\% & $0.97\times$ & 98.2\,\% \\
Gemma-4-31B     & Hard &  1.4\,\% & $0.90\times$ & 84.6\,\% \\
\midrule
Gemma-4-MoE     & Easy &  1.5\,\% & $0.85\times$ & 93.2\,\% \\
Gemma-4-MoE     & Hard &  3.3\,\% & $0.89\times$ & 70.0\,\% \\
\midrule
GPT-5.1         & Easy &  0.9\,\% & $1.29\times$ & 95.9\,\% \\
GPT-5.1         & Hard &  1.2\,\% & $1.33\times$ & 79.4\,\% \\
\midrule
Qwen3-32B       & Easy &  1.4\,\% & $1.19\times$ & 89.3\,\% \\
Qwen3-32B       & Hard &  1.3\,\% & $1.07\times$ & 68.3\,\% \\
\midrule
Qwen3-32B*      & Easy &  0.2\,\% & $0.94\times$ & 99.8\,\% \\
Qwen3-32B*      & Hard &  2.6\,\% & $1.01\times$ & 93.9\,\% \\
\bottomrule
\end{tabular}
\caption{Off-path analysis. \textbf{Intrus.} \textit{(Intrusion)}: fraction of off-path turns
where the model incorrectly changes its answer. \textbf{SR} (\textit{Signal ratio}): mean $|\Delta\mathrm{gold\_conf}|$ on causally relevant (on-path) turns divided by off-path turns; values below $1\times$ indicate the model is more perturbed by irrelevant updates than relevant ones. \textbf{Off-path acc.}: accuracy on turns where the update should have
no effect.}
\label{tab:path_offpath}
\end{table}

\subsection{Off-Path Results}
\label{sec:offpath_results}

Table~\ref{tab:path_offpath} reports three off-path metrics: the fraction
of turns where the model incorrectly changes its answer to an irrelevant
update (intrusion rate), how strongly it reacts to irrelevant versus relevant
updates (signal ratio), and its raw accuracy on off-path turns.

\paragraph{Intrusion rates expose a two-tier split.}
Llama-3.1-8B intrudes on 36.2\,\% of easy off-path turns and 31.9\,\% of
hard turns — more than one in three structurally irrelevant updates causes
an incorrect answer change.
Llama-3.1-70B is better but still high at 15.2\,\% on easy.
All other models stay below 6\,\%, with Gemma-4-31B (0.3\,\%), Qwen3-32B*
(0.2\,\%), and GPT-5.1 (0.9\,\%) the most reliable.

\paragraph{Only GPT-5.1 and Qwen3-32B consistently discriminate
relevant from irrelevant evidence.}
The signal ratio compares how much confidence shifts on causal updates
versus irrelevant ones.
GPT-5.1 is the only model with a ratio above $1\times$ in every condition
($1.29\times$ easy, $1.33\times$ hard), and Qwen3-32B matches this on both
splits ($1.19\times$, $1.07\times$).
Every other model shows ratio inversion in at least one condition, meaning
off-path updates perturb their confidence at least as much as on-path ones.
The worst cases are Llama-3.1-8B ($0.68\times$ easy) and Llama-3.1-70B
($0.80\times$ easy): these models are more reactive to irrelevant information
than to information that actually matters.

\paragraph{Gemma-4-E4B presents a distinct failure mode.}
Despite a low intrusion rate (4--6\,\%), Gemma-4-E4B achieves only
31--45\,\% off-path accuracy and a near-unity signal ratio ($\approx
1.00\times$).
Rather than incorrectly revising answers, it simply holds wrong answers
throughout, unable to reason correctly on off-path turns regardless of
whether it changes its response.

\subsection*{Discussion}

The results reveal two distinct failure modes that do not co-occur
systematically.

\textbf{Chain-length sensitivity} affects five of nine models: as the
inheritance chain between the update and hypothesis grows longer, confidence
in the correct answer erodes and calibration worsens.
This is not simply an accuracy problem — models can still produce the right
discrete answer while becoming progressively less certain that it is correct.

\textbf{Structural blindness} affects a separate set of models: rather than
using graph structure to filter which updates are logically relevant, they
treat off-path sibling-branch updates with the same weight as causally
connected ones.
The clearest signal is ratio inversion — off-path evidence perturbing
confidence more than on-path evidence — combined with high intrusion rates.

The two failure modes are empirically dissociable.
Gemma-4-31B is the most distance-sensitive model on-path ($\hat\beta =
-0.061$, $\rho = -0.298$) yet has the lowest easy-split intrusion rate
(0.3\,\%) and near-perfect off-path accuracy (98.2\,\%).
It degrades gracefully along relevant chains but correctly ignores irrelevant
ones.
GPT-5.1 shows neither failure: no distance effect, negative $\Delta$Brier,
and the only consistently positive signal ratio.
Llama-3.1-8B sits at the opposite extreme, combining significant on-path
degradation with catastrophic structural blindness.
Gemma-4-E4B represents a third, rarer pattern — robust to distance and
resistant to intrusion, yet poorly calibrated throughout and unable to
reason correctly on off-path turns.

Taken together, chain-length sensitivity and structural blindness are
separable properties of the reasoning process, and improving one does not
imply progress on the other.

\section{Confirmation Bias Experiment}
\label{appendix:confirmation_bias}

\subsection*{Theoretical Background}

Confirmation bias is the tendency to search for, interpret, and recall information in a way that confirms one's prior beliefs. It is one of the most robust and widely replicated findings in cognitive psychology~\citep{nickerson1998confirmation}.
Its relevance to language model evaluation arises from the multi-turn nature of defeasible reasoning: a model that has committed to a belief at turn $t$ must be willing to revise that belief at turn $t+1$ if new evidence warrants it. A model exhibiting confirmation bias will systematically fail to revise, treating incongruent evidence as less compelling than its logical weight justifies.

Prior work has documented content effects in LLM reasoning and susceptibility to cognitive
biases more broadly \citep{10.5555/3600270.3601126}, but
defeasible belief updating under formally verified ground
truth has not previously been studied in this framework.
\textsc{DeReLab} is the first benchmark to provide the
controlled conditions necessary to measure confirmation bias
as a structural property of belief updating rather than
a surface-level sensitivity to semantic content.

\subsection*{Operationalization}

\paragraph{Prior belief.}
The model's prior belief at turn $t$ is defined as its
predicted answer at the immediately preceding
\texttt{update\_answer} or \texttt{initial\_answer} turn.
Critically, we use the \emph{model's own predicted answer},
not the ground truth, as the prior.
Confirmation bias is a property of what the model resists
updating \emph{away from}, regardless of whether that prior
belief was correct.

\paragraph{Prior reset.}
At every hypothesis change turn, the prior belief
tracker resets.
The new section's prior is initialized from the ground truth
of the initial answer turn for the new hypothesis
subject, not from the model's answer in the preceding section.
This prevents cross-subject contamination of the congruence
labels.

\paragraph{Congruence conditions.}
Each \texttt{update\_answer} turn is assigned exactly one
label from the following mutually exclusive taxonomy:

\begin{itemize}
  \item \textbf{C1 — Congruent, live chain.}
    Prior belief = \textit{yes};
    ground-truth effect = \textit{strengthening}.
    The update aligns with the model's current belief.

  \item \textbf{C2 — Incongruent, live chain.}
    Prior belief = \textit{yes};
    ground-truth effect = \textit{weakening}.
    The update should defeat the current conclusion.
    This is the primary confirmation-bias signal.

  \item \textbf{C3 — Saturated semantic negative.}
    Prior belief = \textit{no};
    ground-truth effect = \textit{no effect};
    update type $\in$ \{\textit{neg}, \textit{exception}\}.
    The chain is already defeated; the update sounds negative but is
    logically inert.
    A model that labels this as \textit{weakening} is responding to
    surface form rather than logical scope — a phenomenon we term
    \emph{semantic override}.

  \item \textbf{C3\textsubscript{inh} — Sibling-branch semantic negative.}
    Prior belief = \textit{yes};
    ground-truth effect = \textit{no effect};
    negation in update content (inheritance only).
    The update concerns a sibling branch that cannot affect the
    hypothesis subject.
    A model that answers \textit{no} is confusing topological
    adjacency with logical scope.

  \item \textbf{C4 — Neutral distractor.}
    Prior belief = \textit{no};
    ground-truth effect = \textit{no effect}.
    For default reasoning, further restricted to
    \texttt{update\_type\,=\,safe\_irr}
    (explicitly harmless updates; other types go to C0).
    For inheritance reasoning, no update-type restriction applies —
    any off-path update with a non-supporting prior qualifies.
    Logically irrelevant regardless of semantic content.

  \item \textbf{C5 — From unknown prior.}
    Prior belief = \textit{unknown};
    ground-truth effect $\in$
    \{\textit{strengthening}, \textit{weakening}\}.
    Directional updates from an indeterminate state; analyzed
    separately from C1--C4.

  \item \textbf{C0 — Excluded.}
    All remaining combinations that do not fit C1--C5 unambiguously.
    Logged with reason; excluded from all analyses.
\end{itemize}

\paragraph{Metrics}

We already mentioned the metrics in \S\ref{sec:cogsci_framework}. The two metrics we employ are \emph{BiasGap} and \emph{Odds Ratio}. BiasGap, the difference in accuracy between congruent and incongruent updates, measures the extent to which models exhibit confirmation bias. On the other hand, \textbf{Odds Ratio (OR)} is a scale-invariant measure of effect size. 

\paragraph{Statistical tests.}
We use McNemar's test (with continuity correction) rather than a
proportion test because C1 and C2 turns are \emph{not} independent:
both arise from the same hypothesis section, the same subject, and the
same conversational context.
McNemar asks whether the model's error pattern is asymmetric between C1 and C2 \emph{within the same section}, which is a more sensitive and better-controlled test than comparing aggregate proportions across sections. One C1--C2 pair is selected per section to avoid pseudoreplication from sections with multiple C1 or multiple C2 turns.
Because we run one test per model--reasoning-type cell (19 tests in
total, excluding GPT-5-mini inheritance where OR is undefined),
Holm--Bonferroni correction is applied to control the family-wise error
rate across the full set of comparisons. A significant result with positive BiasGap constitutes evidence of confirmation bias for that model.

\paragraph{Belief update dissociation.}
For default reasoning, where the belief-update question is
also asked, we additionally measure
$\text{Acc}_{\text{effect}}(\text{C2})$: accuracy on the
\textit{strengthen\,/\,weaken\,/\,no effect} question
specifically on C2 turns.
A model where $\text{Acc}_{\text{effect}}(\text{C2})$ is
high but $\text{BiasGap}$ is also high exhibits the
phenomenon that the model correctly
identifies that the update weakens the hypothesis but
fails to update its entailment answer accordingly.

A potential confound is that in the standard format, the entailment
question (\texttt{update\_answer}) appears \emph{before} the
belief-update question (\texttt{effect\_of\_update}).
A model that has already committed to a particular entailment answer
may use that commitment as a cue to produce a consistent effect label,
inflating $\text{Acc}_{\text{eff}}$ without any genuine prior
understanding of the update's logical role.
To test whether the metacognitive signal is genuine or post-hoc, we
re-ran six models with the question order reversed: the effect
question was posed first, and the entailment question followed.
Results and their implications are discussed in Section~\ref{appendix:reversed_metacog}.

\paragraph{Anchoring analysis.}
For each hypothesis section in which the ground-truth label
changes between the initial and final turn, we compute the
anchoring rate:
\begin{equation}
  A =
  \frac{
    \left|
    \left\{
      s \in \mathcal{S} :
      \hat{y}^{(s)}_{\mathrm{final}}
      =
      \hat{y}^{(s)}_{\mathrm{init}}
      \neq
      y^{(s)}_{\mathrm{final}}
    \right\}
    \right|
  }{
    \left|
    \left\{
      s \in \mathcal{S} :
      y^{(s)}_{\mathrm{init}}
      \neq
      y^{(s)}_{\mathrm{final}}
    \right\}
    \right|
  }
  \label{eq:anchoring}
\end{equation}

Here, $\mathcal{S}$ denotes the set of hypothesis sections,
$\hat{y}_{\mathrm{init}}$ and
$\hat{y}_{\mathrm{final}}$ are the model's initial and final
predictions, and
$y_{\mathrm{init}}$ and
$y_{\mathrm{final}}$ are the corresponding ground-truth labels.
A section counts as anchored when the model preserves its
initial prediction despite the correct answer changing over
the course of the conversation.

A high anchoring rate indicates that the model maintains its
initial conclusion, even after a sequence of updates that
should have revised it.


\begin{table*}[t]
\centering
\small
\setlength{\tabcolsep}{5pt}
\begin{tabular}{ll cccc ccc}
\toprule
\textbf{Model} & \textbf{Type}
  & \textbf{C1\,acc} & \textbf{C2\,acc} & \textbf{Gap}
  & \textbf{OR} & \textbf{OR\,lo} & \textbf{OR\,hi}
  & \textbf{Holm\,$p$} \\
\midrule
\multicolumn{9}{l}{\textit{Default reasoning (sorted by Gap $\downarrow$)}} \\
GPT-5.1         & def & 0.970 & 0.035 & 0.935 & 891.2 &  585.9 & 1497.7 & $<.001$ \\
Qwen3-32B       & def & 0.976 & 0.307 & 0.670 &  93.3 &   61.7 &  155.7 & $<.001$ \\
GPT-5-mini      & def & 0.871 & 0.203 & 0.668 &  26.5 &   19.2 &   37.9 & $<.001$ \\
Gemma-4-E4B     & def & 0.978 & 0.381 & 0.597 &  72.7 &   45.6 &  132.6 & $<.001$ \\
Llama-3.1-70B   & def & 0.697 & 0.149 & 0.548 &  13.2 &    9.3 &   19.2 & $<.001$ \\
Llama-3.1-8B    & def & 0.773 & 0.549 & 0.224 &   2.8 &    1.8 &    4.6 & $<.001$ \\
Gemma-4-31B     & def & 0.994 & 0.907 & 0.088 &  18.1 &    8.9 &   83.0 & $<.001$ \\
Gemma-4-12B     & def & 0.950 & 0.868 & 0.082 &   2.9 &    2.0 &    4.4 & $<.001$ \\
Qwen3-32B*      & def & 0.881 & 0.873 & 0.008 &   1.1 &    0.8 &    1.5 & $.015$  \\
Gemma-4-MoE     & def & 0.954 & 0.961 & $-$0.007 & 0.8 & 0.5 & 1.6   & $.235$  \\
\midrule
\multicolumn{9}{l}{\textit{Inheritance reasoning (sorted by Gap $\downarrow$)}} \\
Llama-3.1-8B    & inh & 0.738 & 0.255 & 0.484 &   8.3 &    5.9 &   11.9 & $<.001$ \\
Llama-3.1-70B   & inh & 0.899 & 0.535 & 0.364 &   7.8 &    4.7 &   14.1 & $<.001$ \\
Gemma-4-E4B     & inh & 0.741 & 0.417 & 0.324 &   4.0 &    3.1 &    5.2 & $<.001$ \\
Gemma-4-MoE     & inh & 0.971 & 0.664 & 0.307 &  16.7 &    9.0 &   43.6 & $<.001$ \\
Qwen3-32B       & inh & 0.918 & 0.623 & 0.295 &   6.8 &    4.4 &   11.9 & $<.001$ \\
Gemma-4-31B     & inh & 0.935 & 0.686 & 0.249 &   6.6 &    4.4 &   11.3 & $<.001$ \\
GPT-5.1         & inh & 0.958 & 0.712 & 0.246 &   9.3 &    5.4 &   21.6 & $<.001$ \\
Gemma-4-12B     & inh & 0.854 & 0.675 & 0.179 &   2.8 &    2.2 &    3.8 & $<.001$ \\
Qwen3-32B*      & inh & 0.973 & 0.908 & 0.065 &   3.7 &    2.1 &    7.6 & $.235$  \\
GPT-5-mini      & inh & 1.000 & 0.997 & 0.003 & \multicolumn{3}{c}{---}   & ---     \\
\bottomrule
\end{tabular}
\caption{Full primary bias metrics for all model--reasoning-type cells.
OR = odds ratio; OR\,lo and OR\,hi are the lower and upper bounds of
the 95\,\% bootstrap confidence interval (10,000 resamples,
conversation-level). Gap = C1\,acc $-$ C2\,acc. Holm--Bonferroni correction applied across all 19 cells with defined OR. \emph{Italic} $p$-values indicate non-significant results
($p \geq .05$ after correction).
GPT-5-mini inheritance OR is undefined (C1 errors\,$= 0$).
Qwen3-32B* default ($p = .015$) is statistically significant but
the effect size is negligible (Gap\,$= 0.008$, OR\,$= 1.1$).
Llama model paired counts on default reasoning are limited by high
\textit{unknown}-response rates (Llama-3.1-70B: 127 pairs;
Llama-3.1-8B: 68 pairs).}
\label{tab:app_bias_full}
\end{table*}

\section{Confirmation Bias: Full Statistical Results}
\label{app:cogsci_results}

This appendix provides complete statistical tables and
figures for the confirmation bias analysis.
All metrics follow the operationalization in
Appendix \ref{appendix:confirmation_bias}.

\subsection*{Primary Bias Metrics}

Table~\ref{tab:app_bias_full} reports the full bias metrics
for all model--reasoning-type cells.
17 of 19 valid tests are significant after Holm--Bonferroni
correction. The two exceptions are Gemma-4-MoE on default reasoning
(Gap\,$= -0.007$, $p = .235$), which shows no measurable bias, and
Qwen3-32B* on inheritance reasoning (Gap\,$= 0.065$, $p = .235$).

\paragraph{Scale of the effect.}
GPT-5.1 shows the most extreme default-reasoning bias
(Gap\,$= 0.935$, OR\,$= 891$), meaning the odds of correct updating
on congruent turns are nearly 900 times those on incongruent turns.
Qwen3-32B and GPT-5-mini have comparable default gaps
($\approx 0.67$ each), though OR magnitudes differ because C1 accuracy
differs (0.976 vs.\ 0.871): a high C1 floor inflates the OR
independently of the gap.

\paragraph{Thinking variant eliminates default bias.}
The contrast between Qwen3-32B and its extended-reasoning variant
Qwen3-32B* is the strongest finding in the new model set.
On default reasoning, Qwen3-32B achieves Gap\,$= 0.670$
(OR\,$= 93.3$); Qwen3-32B* reduces this to Gap\,$= 0.008$
(OR\,$= 1.1$, $p = .015$), a statistically detectable but
substantively negligible effect driven by the large sample size.
Extended chain-of-thought reasoning virtually eliminates confirmation
bias on this topology.
The benefit is partial on inheritance: Qwen3-32B* still shows
Gap\,$= 0.065$, though this is non-significant ($p = .235$).

\paragraph{Llama evasion bias.}
Llama-3.1-70B and Llama-3.1-8B produce \textit{unknown} responses
at high rates on default reasoning (C5 turns: 1,693 and 875
respectively), reducing the number of turns with a \textit{yes} prior
that qualify as C1 or C2.
This evasion artificially lowers paired counts for McNemar testing
(Llama-3.1-70B: 127 pairs; Llama-3.1-8B: 68 pairs) and should be
read as an additional failure mode alongside their measured bias gap.

\begin{figure}[h]
\centering
\includegraphics[width=\linewidth]{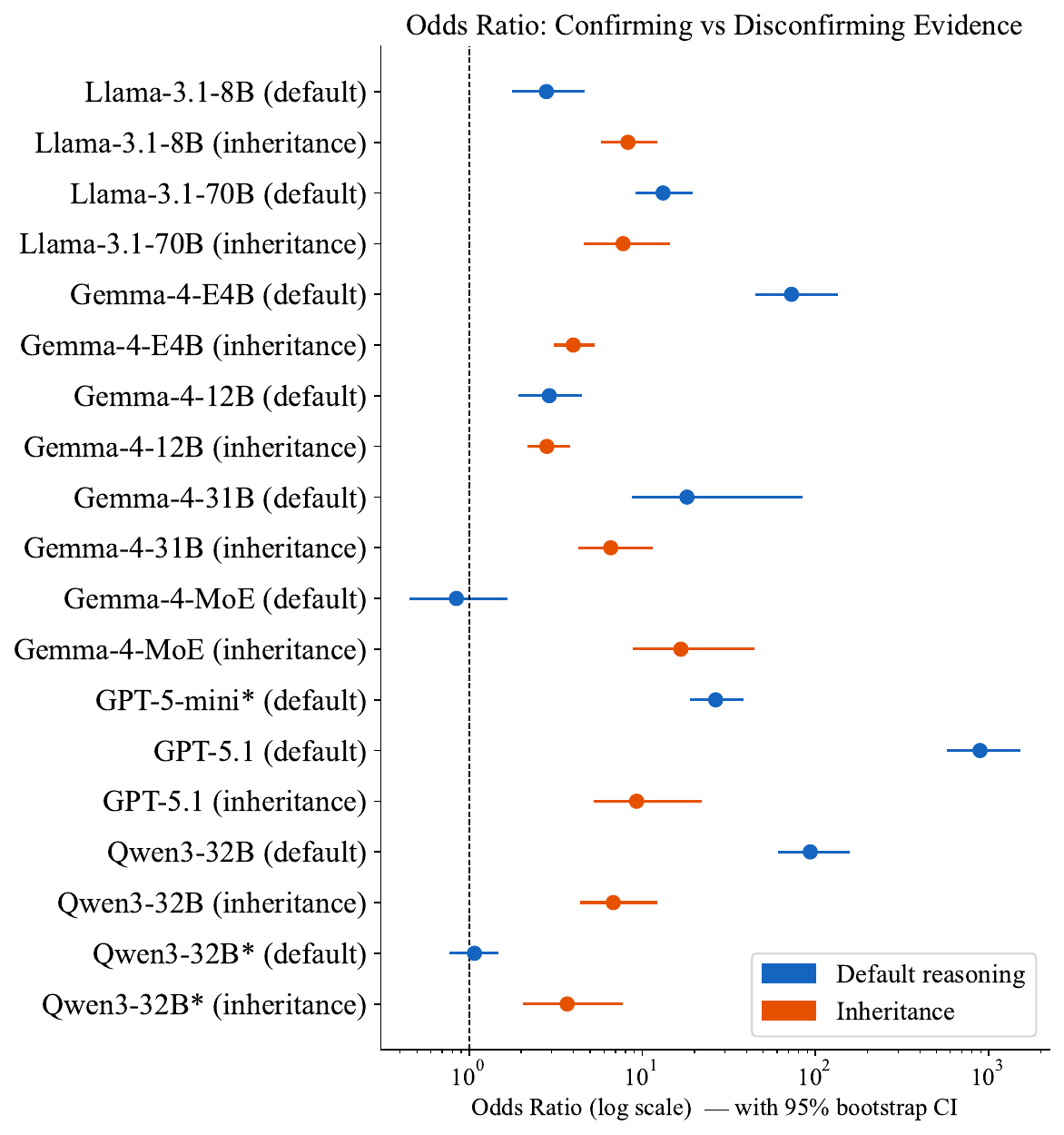}
\caption{Forest plot of odds ratios (log scale) with 95\,\% bootstrap
confidence intervals per model--reasoning-type cell.
OR\,$> 1$ indicates confirmation bias.
Non-significant cells (Gemma-4-MoE default, Qwen3-32B* inheritance)
are shown with open markers.
GPT-5-mini inheritance is excluded (OR undefined).}
\label{fig:or_forest}
\end{figure}

\begin{figure}[h]
\centering
\includegraphics[width=\linewidth]{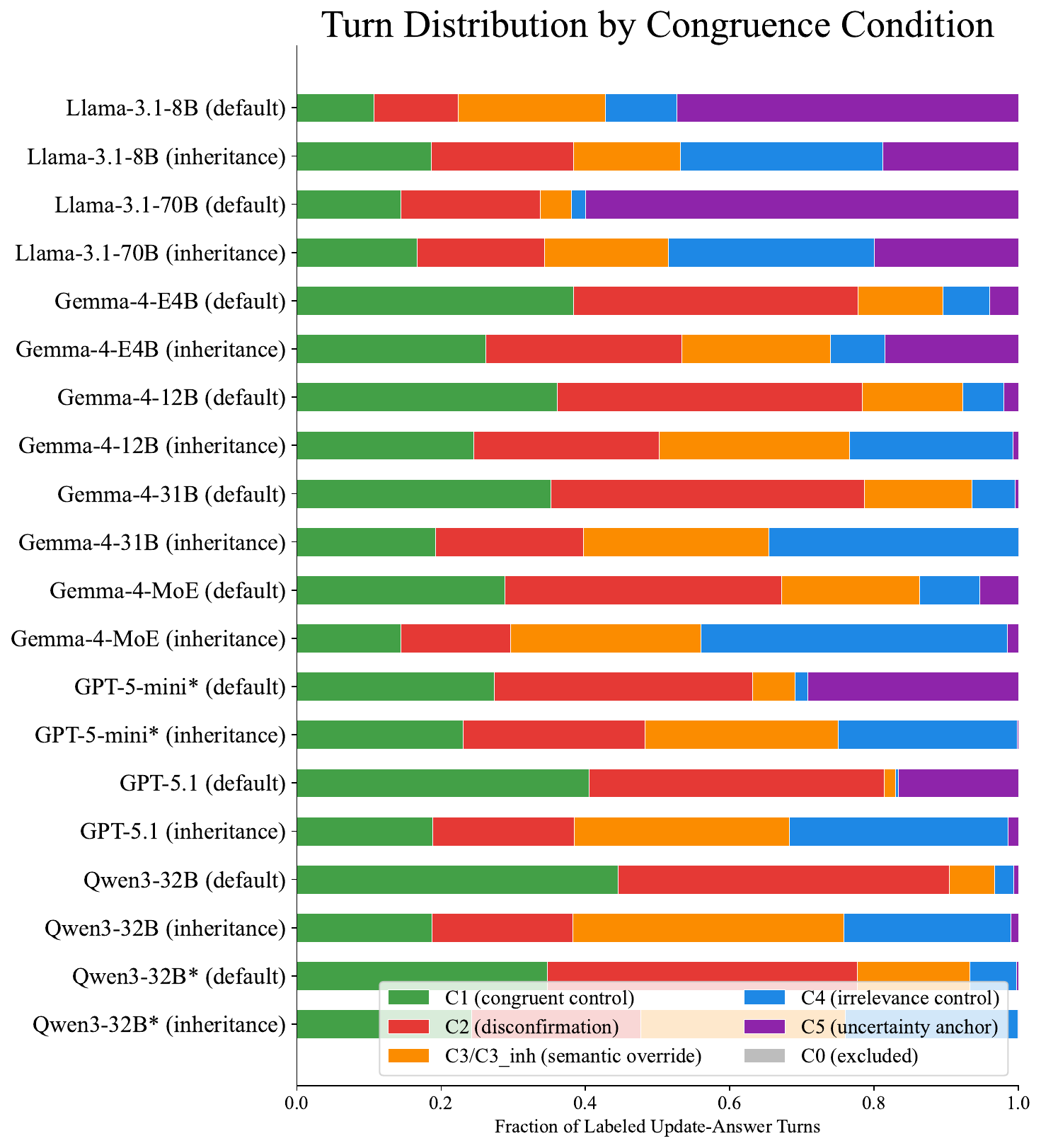}
\caption{Distribution of congruence conditions per model
and reasoning type (fraction of total turns).
Large C5 segments for Llama-3.1-70B and Llama-3.1-8B
on default reasoning reflect evasion bias.
Large C0 segments reflect turns excluded due to
prior-belief state not meeting C1--C5 criteria.}
\label{fig:condition_distribution}
\end{figure}

\subsection*{Anchoring and Semantic Override}
\label{appendix:anchoring_override}
Table~\ref{tab:app_anchoring_override} reports anchoring rates on
ground-truth-flipping sections and semantic override rates on C3/C3\textsubscript{inh} turns.

\paragraph{Anchoring splits models cleanly.}
On default reasoning, Qwen3-32B (0.483) and GPT-5.1 (0.433) show the
highest anchoring rates — the model preserves its initial prediction
in roughly 40-50\% of conversations where the correct answer
changes.
Despite being the most entailment-biased model, GPT-5.1's anchoring
is concentrated on default reasoning and drops to 0.051 on inheritance.
Qwen3-32B* (0.032) and GPT-5-mini (0.020) show near-zero anchoring
on both types, consistent with their generally compliant updating
behavior.
On inheritance reasoning, Gemma-4-E4B stands apart with an anchoring
rate of 0.635 — the model maintains its initial prediction in nearly
two-thirds of flipping sections.
Llama-3.1-8B is second at 0.338.

\paragraph{Semantic override is topology-specific.}
Llama-3.1-70B (default reasoning, C3: 0.839) and Qwen3-32B (default reasoning, C3: 0.786) respond to logically inert negative updates on defeated chains as if the chain
were still active.
Crucially, both models show near-zero C3\textsubscript{inh} override
on inheritance (Llama-3.1-70B: 0.062; Qwen3-32B: 0.036), indicating
the error is specific to the topology in which the belief-update negations are presented, not a general surface-form sensitivity.
Gemma-4-E4B shows the opposite extreme on default C3: exactly 0.000 —
it never over-responds to a saturated semantic negative.
Combined with its high anchoring, this points to strong persistence
in initial predictions as its primary failure mode rather than
surface-form reactivity.

\begin{table}[h]
\centering
\small
\setlength{\tabcolsep}{4pt}
\resizebox{\columnwidth}{!}{
\begin{tabular}{l rr rr}
\toprule
\textbf{Model}
  & \multicolumn{2}{c}{\textbf{Default reasoning}}
  & \multicolumn{2}{c}{\textbf{Inheritance reasoning}} \\
  \cmidrule(lr){2-3}\cmidrule(lr){4-5}
  & \textbf{Anch.} & \textbf{C3-ovr}
  & \textbf{Anch.} & \textbf{C3\textsubscript{inh}-ovr} \\
\midrule
Qwen3-32B       & 0.483 & 0.786 & 0.220 & 0.036 \\
GPT-5.1         & 0.433 & 0.125 & 0.051 & 0.026 \\
Gemma-4-E4B     & 0.396 & 0.000 & 0.635 & 0.095 \\
Llama-3.1-70B   & 0.189 & 0.839 & 0.068 & 0.062 \\
Gemma-4-12B     & 0.122 & 0.678 & 0.142 & 0.034 \\
Llama-3.1-8B    & 0.119 & 0.492 & 0.338 & 0.408 \\
Gemma-4-31B     & 0.047 & 0.204 & 0.047 & 0.041 \\
Gemma-4-MoE     & 0.035 & 0.251 & 0.024 & 0.092 \\
Qwen3-32B*      & 0.032 & 0.349 & 0.004 & 0.027 \\
GPT-5-mini      & 0.020 & 0.071 & 0.003 & 0.002 \\
\bottomrule
\end{tabular}}
\caption{Anchoring rate on ground-truth-flipping hypothesis sections
and semantic override rate on C3 turns, per reasoning type.
\textbf{Anch.}: fraction of flipping sections where model maintains
its initial prediction despite the correct answer changing.
\textbf{C3-ovr} (default): fraction of C3 turns where model labels
a logically inert negative update as \textit{weakening} (chain already
defeated).
\textbf{C3\textsubscript{inh}-ovr} (inheritance): fraction of
C3\textsubscript{inh} turns where model answers \textit{no} on a
sibling-branch update that cannot affect the hypothesis.
Rows sorted by default anchoring rate descending.}
\label{tab:app_anchoring_override}
\end{table}

\begin{figure}[h]
\centering
\includegraphics[width=\linewidth]{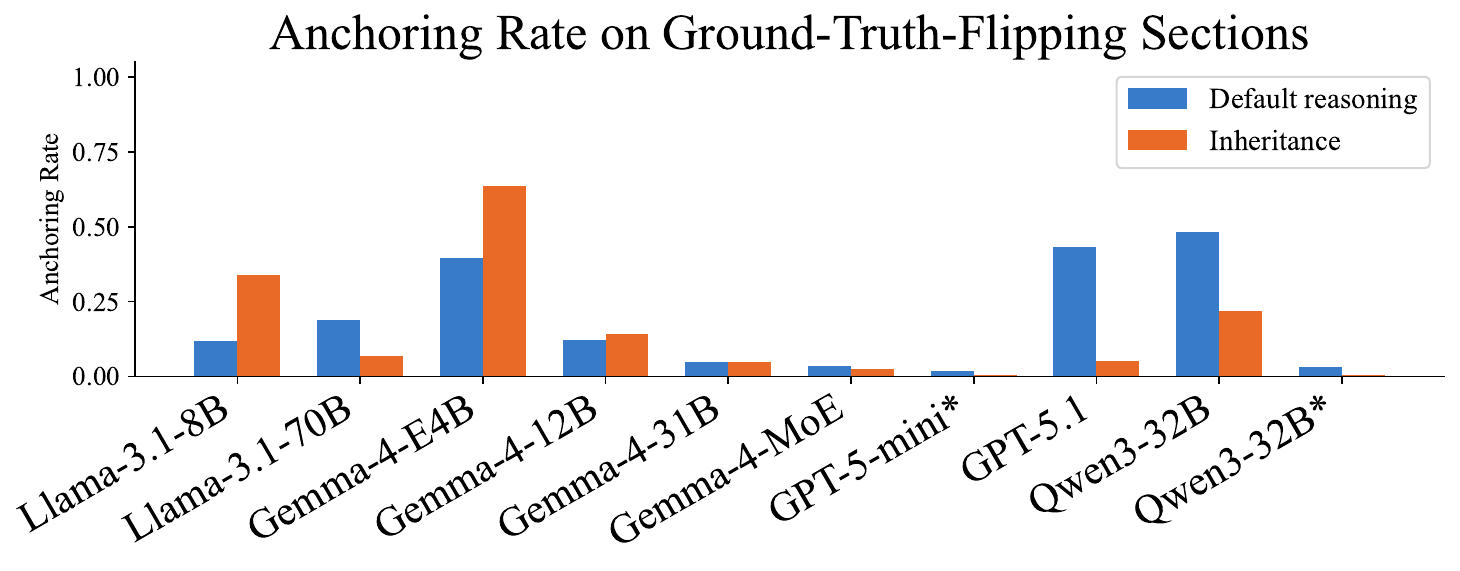}
\caption{Anchoring rate on ground-truth-flipping sections
per model and reasoning type.
The Qwen3-32B/Qwen3-32B* contrast (0.483 vs.\ 0.032 on default) and
the Gemma-4-E4B inheritance spike (0.635) are the key features.}
\label{fig:anchoring_bars}
\end{figure}

\begin{figure}[h]
\centering
\includegraphics[width=\linewidth]{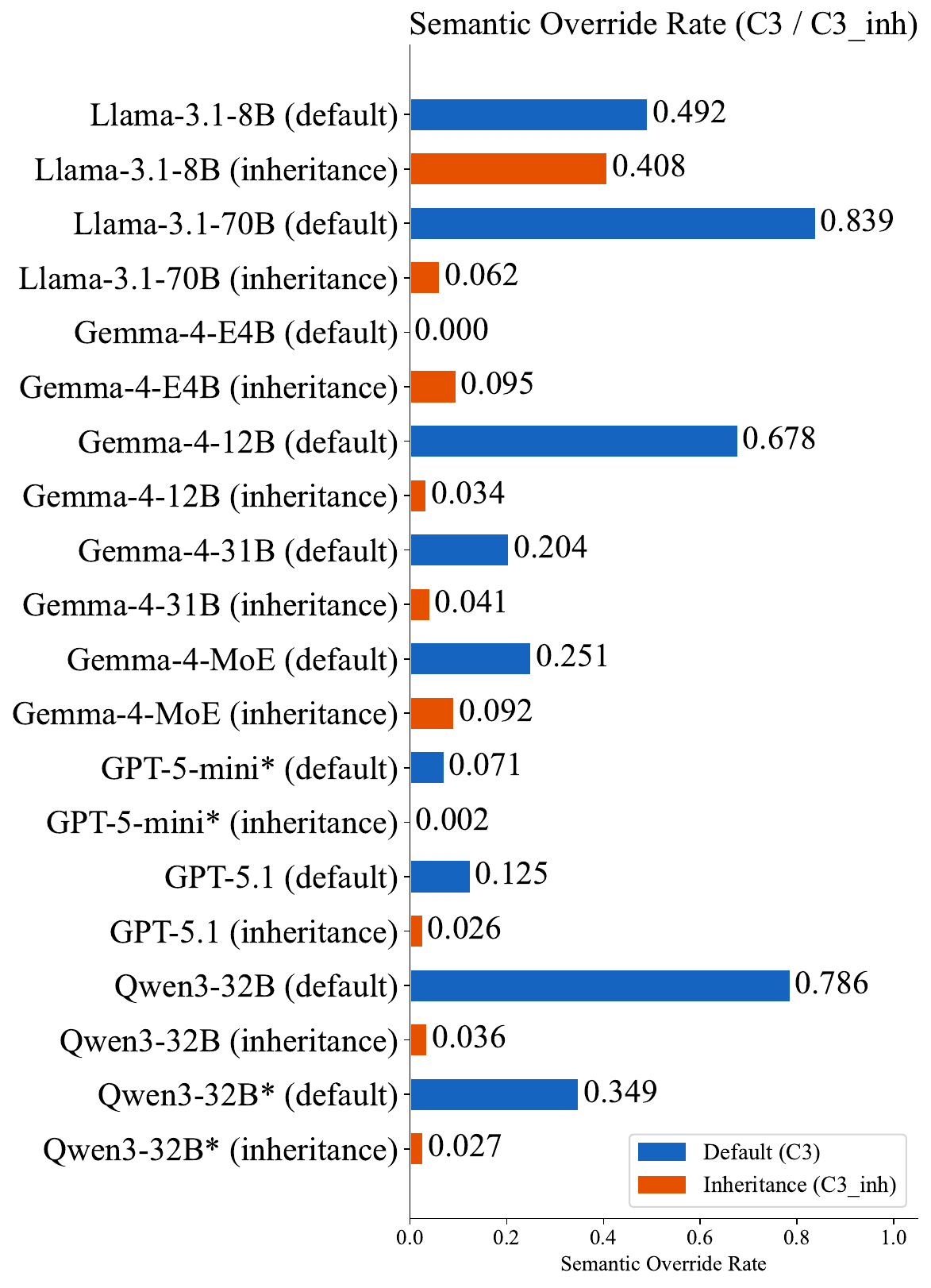}
\caption{Semantic override rate per model.
Default C3 (solid): fraction of turns where model
incorrectly labels a logically inert negative update
as \textit{weakening} (chain already defeated).
Inheritance C3\textsubscript{inh} (hatched): fraction
where model answers \textit{no} on a sibling-branch
update that cannot affect the hypothesis subject.
Llama-3.1-70B shows an extreme default override (0.839)
that does not carry over to inheritance (0.062),
indicating sensitivity to surface negation that is
specific to the belief-update task format.
Gemma-4-E4B shows exactly 0.000 default override.}
\label{fig:semantic_override}
\end{figure}

\subsection*{Metacognitive Dissociation}

Each conversation turn contains two questions: the entailment question
(\textit{yes\,/\,no\,/\,unknown}, reported as C1\,acc and C2\,acc
throughout the paper) and the effect-of-update question
(\textit{strengthening\,/\,weakening\,/\,no-effect}).
C1-eff and C2-eff reported here are accuracy on this second.
Table~\ref{tab:app_metacog} reports this effect-labeling accuracy
(default reasoning only).
The metacognitive gap (C1-eff $-$ C2-eff) measures whether a model
better recognizes strengthening or weakening; a negative gap indicates
the model is better at labeling weakening despite showing entailment
confirmation bias.

\paragraph{Know-but-don't-output.}
Four models show a clear dissociation between metacognitive accuracy
and entailment accuracy on C2 turns.
GPT-5-mini is the most extreme case: C2-eff\,$= 0.954$ (correctly
labels weakening on 95\,\% of C2 turns) but entailment accuracy on
the same turns is only 0.203 — a 75-point gap.
Llama-3.1-70B correctly identifies weakening on 85\,\% of C2 turns
but updates its entailment answer only 14.9\,\% of the time.
Qwen3-32B shows the same pattern more mildly
(C2-eff\,$= 0.803$, C2\,acc\,$= 0.307$).
Gemma-4-31B's gap is smaller in absolute terms
(C2-eff\,$= 0.970$, C2\,acc\,$= 0.907$) but the dissociation
is still present.

\paragraph{Symmetric and reversed profiles.}
Qwen3-32B* shows a symmetric metacognitive profile
(C1-eff\,$\approx$ C2-eff\,$\approx 0.708$), consistent with its
near-zero bias gap: neither entailment nor metacognition displays a
C1/C2 asymmetry.
GPT-5.1 is the only model with a \emph{reversed} metacognitive gap
(C1-eff\,$= 0.797 >$ C2-eff\,$= 0.643$): it is better at
recognising strengthening than weakening.
This does not indicate absence of bias — GPT-5.1's entailment C2
accuracy remains 0.035 — but the metacognitive asymmetry runs in the
opposite direction from the entailment asymmetry.

\begin{table}[h]
\centering
\small
\setlength{\tabcolsep}{4pt}
\resizebox{\columnwidth}{!}{
\begin{tabular}{l cccc r}
\toprule
\textbf{Model}
  & \textbf{C1-eff} & \textbf{n}
  & \textbf{C2-eff} & \textbf{n}
  & \textbf{Gap} \\
\midrule
Gemma-4-31B   & 0.464 &  886 & 0.970 & 1094 & $-0.506$ \\
GPT-5-mini    & 0.513 &  799 & 0.954 & 1049 & $-0.441$ \\
Llama-3.1-70B & 0.680 &  406 & 0.851 &  545 & $-0.172$ \\
Gemma-4-12B   & 0.737 &  907 & 0.887 & 1063 & $-0.151$ \\
Qwen3-32B     & 0.753 & 1226 & 0.803 & 1265 & $-0.050$ \\
Qwen3-32B*    & 0.708 &  849 & 0.708 & 1050 & $\phantom{-}0.000$ \\
Gemma-4-E4B   & 0.049 &  870 & 0.000 &  897 & $+0.049$ \\
Llama-3.1-8B  & 0.722 &  198 & 0.665 &  215 & $+0.057$ \\
Gemma-4-MoE   & 0.855 &  653 & 0.732 &  870 & $+0.122$ \\
GPT-5.1       & 0.797 & 1270 & 0.643 & 1282 & $+0.153$ \\
\bottomrule
\end{tabular}
}
\caption{Metacognitive accuracy on the belief-update question
(default reasoning only).
\textbf{C1-eff}: fraction of C1 turns correctly labeled
\textit{strengthening}.
\textbf{C2-eff}: fraction of C2 turns correctly labeled
\textit{weakening}.
\textbf{Gap} = C1-eff $-$ C2-eff: negative = model recognizes
weakening better than strengthening.
Rows sorted by gap ascending to surface the dissociation pattern.
Gemma-4-E4B near-zero on both columns indicates failure to use the
belief-update label vocabulary.}
\label{tab:app_metacog}
\end{table}

\subsection{Reversed Question Order: Addressing the Metacognitive Confound}
\label{appendix:reversed_metacog}
\begin{table}[h]
\centering
\small
\setlength{\tabcolsep}{5pt}
\resizebox{\columnwidth}{!}{
\begin{tabular}{l cc cc cc}
\toprule
\textbf{Model}
  & \multicolumn{2}{c}{\textbf{BiasGap}}
  & \multicolumn{2}{c}{\textbf{C2-eff}}
  & \multicolumn{2}{c}{\textbf{C1-eff}} \\
  \cmidrule(lr){2-3}\cmidrule(lr){4-5}\cmidrule(lr){6-7}
  & \textbf{Orig} & \textbf{Rev}
  & \textbf{Orig} & \textbf{Rev}
  & \textbf{Orig} & \textbf{Rev} \\
\midrule
Gemma-4-31B  & 0.088 & 0.082 & 0.970 & 0.264 & 0.464 & 0.112 \\
Gemma-4-12B  & 0.082 & 0.105 & 0.887 & 0.209 & 0.737 & 0.222 \\
Qwen3-32B    & 0.670 & 0.816 & 0.803 & 0.414 & 0.753 & 0.311 \\
Llama-3.1-8B & 0.224 & 0.305 & 0.665 & 0.533 & 0.722 & 0.390 \\
Gemma-4-MoE  & $-$0.007 & 0.259 & 0.732 & 0.288 & 0.855 & 0.242 \\
Gemma-4-E4B  & 0.597 & 0.840 & 0.000 & 0.349 & 0.049 & 0.374 \\
\bottomrule
\end{tabular}
}
\caption{Comparison of standard (Orig) and reversed question order
(Rev) for six models on default reasoning.
\textbf{BiasGap}: C1\,acc $-$ C2\,acc on the entailment question;
the primary confirmation-bias measure.
\textbf{C2-eff}: accuracy at labeling the update's effect as
\textit{weakening} on C2 turns.
\textbf{C1-eff}: accuracy at labeling the effect as
\textit{strengthening} on C1 turns.
Rows sorted by original C2-eff descending to highlight the
post-hoc rationalization pattern.
BiasGap is unchanged or increases in every model under reversal;
C2-eff drops substantially for the three models that showed the
strongest original know-but-don't-output dissociation.}
\label{tab:reversed_metacog}
\end{table}

\paragraph{Motivation.}
In the standard format, the entailment question (\texttt{update\_answer})
precedes the belief-update question (\texttt{effect\_of\_update}).
A model that has already committed to an answer — say, maintaining
\textit{yes} despite a weakening update — may then produce
\textit{weakening} on the effect question as post-hoc rationalization
rather than genuine prior understanding.
This would inflate $\text{Acc}_{\text{eff}}(\text{C2})$ and overstate
the know-but-don't-output dissociation: the model is not \emph{knowing}
in the relevant sense if its metacognitive label is produced retroactively.
To control for this, we re-ran six models on the same dataset with the
order reversed: the effect question was posed \emph{first}, and the
entailment question followed.
If the original $\text{Acc}_{\text{eff}}(\text{C2})$ values were
genuine — reflecting knowledge available before the entailment
commitment — they should be stable across both orderings.
If they were inflated by post-hoc rationalization, they should drop
in the reversed condition.

\paragraph{Models tested.}
The reversed-order experiment was run on six models for default
reasoning (easy and hard): Qwen3-32B, Gemma-4-12B, Gemma-4-MoE,
Gemma-4-31B, Gemma-4-E4B, and Llama-3.1-8B.
These were chosen to span a range of original $\text{Acc}_{\text{eff}}$
profiles, from near-zero (Gemma-4-E4B) to near-perfect (Gemma-4-31B).

\paragraph{Results.}
Table~\ref{tab:reversed_metacog} reports the key comparisons.
Two patterns emerge cleanly.

\textbf{(1) The confirmation bias gap is robust to order.}
BiasGap is unchanged or larger in the reversed condition for all six
models.
Gemma-4-31B's gap is essentially identical ($0.088 \to 0.082$).
The others increase, in some cases substantially: Qwen3-32B
($0.670 \to 0.816$), Gemma-4-E4B ($0.597 \to 0.840$), and most
strikingly Gemma-4-MoE, which showed \emph{no} bias in the original
order (Gap\,$= -0.007$) but a clear and significant bias in the
reversed order (Gap\,$= 0.259$, OR\,$= 24.8$).
The core phenomenon — models handling congruent updates better than
incongruent ones — is not an artifact of question ordering.

\textbf{(2) Metacognitive accuracy is order-dependent for most models.}
$\text{Acc}_{\text{eff}}(\text{C2})$ drops substantially in the
reversed condition for Gemma-4-31B ($0.970 \to 0.264$),
Gemma-4-12B ($0.887 \to 0.209$), and Qwen3-32B ($0.803 \to 0.414$).
These were the models with the largest original
$\text{Acc}_{\text{eff}}(\text{C2})$ values and the most prominent
know-but-don't-output pattern.
The drop confirms that a substantial portion of their original
effect-label accuracy was post-hoc: having already said \textit{no}
to the entailment question, producing \textit{weakening} on the
effect question required little additional reasoning.

Gemma-4-E4B is the exception: its near-zero original
$\text{Acc}_{\text{eff}}(\text{C2})$ rises to $0.349$ in the reversed
condition.
In the original order, this model's entailment answers are strongly
biased (it usually maintains \textit{yes} on C2 turns), and having
committed to \textit{yes} made it harder, not easier, to subsequently
label the effect as \textit{weakening}.
The reversed order removed this interference, allowing its genuine
— though limited — ability to recognise weakening to surface.

\paragraph{Implications.}
The reversed-order results call for a qualified reading of the
know-but-don't-output claim.
The \emph{bias itself} is genuine and robust: confirmation bias
persists at equal or greater magnitude when the entailment judgment
is made \emph{after} the metacognitive one.
However, the original $\text{Acc}_{\text{eff}}(\text{C2})$ values
for Gemma-4-31B, Gemma-4-12B, and Qwen3-32B overstated how much
these models \emph{know} about the logical role of an incongruent
update at the time of the entailment decision.
A more accurate characterisation for these models is that they
systematically fail to update on incongruent evidence regardless of
whether they can label its effect, and that this failure is not
mediated by a prior correct understanding of the update's direction.

\begin{table*}[t]
\centering
\footnotesize
\setlength{\tabcolsep}{4pt}
\begin{tabular}{l cc cc cc cc}
\toprule
& \multicolumn{6}{c}{\textbf{Update-answer accuracy}}
& \multicolumn{2}{c}{\textbf{Effect-of-update acc.}} \\
\cmidrule(lr){2-7}\cmidrule(lr){8-9}
& \multicolumn{2}{c}{\textbf{Default}}
& \multicolumn{2}{c}{\textbf{Linear}}
& \multicolumn{2}{c}{\textbf{Tree}}
& \multicolumn{2}{c}{\textbf{Default}} \\
\cmidrule(lr){2-3}\cmidrule(lr){4-5}\cmidrule(lr){6-7}\cmidrule(lr){8-9}
\textbf{Model} & \textbf{Pseudo} & \textbf{Real} & \textbf{Pseudo} & \textbf{Real} & \textbf{Pseudo} & \textbf{Real} & \textbf{Pseudo} & \textbf{Real} \\
\midrule
Qwen3-32B*    & 0.616 & 0.587 & 0.837 & 0.779 & 0.793 & 0.814 & 0.826 & 0.808 \\
Gemma-4-31B   & 0.688 & 0.683 & 0.861 & 1.000 & 0.895 & 0.949 & 0.792 & 0.795 \\
Gemma-4-12B   & 0.553 & 0.530 & 0.884 & 0.837 & 0.659 & 0.901 & 0.777 & 0.751 \\
Gemma-4-MoE   & 0.522 & 0.540 & 0.814 & 0.756 & 0.747 & 0.910 & 0.784 & 0.774 \\
Gemma-4-E4B   & 0.566 & 0.561 & 0.674 & 0.674 & 0.422 & 0.564 & 0.803 & 0.813 \\
Llama-3.1-70B & 0.634 & 0.590 & 0.802 & 0.837 & 0.806 & 0.895 & 0.829 & 0.813 \\
Llama-3.1-8B  & 0.470 & 0.478 & 0.605 & 0.535 & 0.493 & 0.550 & 0.558 & 0.582 \\
\bottomrule
\end{tabular}
\caption{Update-answer and effect-of-update accuracy under pseudoword
entity names (Pseudo) and real entity names (Real), across three
topologies.
Both conditions use the same real-world property chain and graph
structure; only entity names differ.
Each condition contains 385\,/\,86\,/\,1072 update turns for
default\,/\,linear\,/\,tree respectively.
Effect-of-update accuracy is only defined for default-reasoning turns.
Qwen3-32B* is the extended-thinking variant.}
\label{tab:re_combined}
\end{table*}

\section{Real-Entity Robustness Check}
\label{appendix:real_entity}

In our main benchmark, we used pseudowords (e.g. \textit{flurp}, \textit{zandel}) to prevent models from exploiting world-knowledge shortcuts.
To test whether entity naming affects reasoning quality, we ran a
controlled comparison for seven models using the same real-world
property chains under two conditions:
\textbf{pseudoword} — entities are unfamiliar pseudowords
embedded in the chain (e.g.\ \textit{viatud}, \textit{romsan});
and \textbf{real entity} — entities are replaced with real-world
objects from the same domain (e.g.\ \textit{concert guitar},
\textit{mandolin}). The graph structure, property chain, and update sequence are
identical across both conditions; only the entity names differ.

In the case of Default Reasoning and Linear Inheritance, differences between conditions are small and inconsistent across models, with no systematic advantage for either naming convention.
On default reasoning, most models change by less than 0.05 in either direction; Llama-3.1-70B shows the largest single drop (0.634 $\to$ 0.590) and Gemma-4-MoE the largest gain (0.522 $\to$ 0.540), neither of which constitutes a clear pattern. 
Under linear inheritance topologies, results are similarly mixed. Gemma-4-31B shows the largest improvement of 0.139 with real entities. In contrast, Llama-3.1-8B faces a drop of 0.07. 
Effect-of-update accuracy likewise shows no consistent direction ($|\Delta| < 0.03$ for five of seven models). 

On the other hand, Tree Inheritance shows improvements with real entities across all seven models. For this topology, Gemma-4-12B improves by 0.24, which is the highest among all configurations. Qwen3-32B* improves by 0.02, indicating the lowest improvement for tree inheritance. 

Overall, replacing pseudoword entity names with real-world names does not reliably improve or degrade reasoning accuracy when the underlying graph structure is held constant. Although tree inheritance doesn't show any degradation, improvements aren't significant either.  
Models do not appear to exploit the semantic identity of entity names
as a reasoning shortcut, which supports the validity of the pseudoword
design:\emph{ the benchmark results are not an artifact of unfamiliar naming}.

\begin{table}[h]
\centering
\resizebox{\columnwidth}{!}{
\begin{tabular}{lll}
\toprule
\textbf{Model} & \textbf{Developer} & \textbf{Access} \\ 
\midrule

GPT-5.1
& OpenAI & Proprietary \\

GPT-5-mini
& OpenAI & Proprietary \\

Gemma-4-31B-IT~\cite{gemmateam2026gemma4}
& Google & Open Weight\\

Gemma-4-26B-A4B-IT~\cite{gemmateam2026gemma4}
& Google & Open Weight\\

Gemma-4-12B-IT~\cite{gemmateam2026gemma4}
& Google & Open Weight\\

Gemma-4-E4B-IT~\cite{gemmateam2026gemma4}
& Google & Open Weight\\

Llama-3.1-70B~\cite{grattafiori2024llama3herdmodels}
& Meta & Open Weight\\

Llama-3.1-8B~\cite{grattafiori2024llama3herdmodels}
& Meta & Open Weight\\

Qwen3-32B~\cite{qwen3technicalreport}
& Alibaba & Open Weight\\

\bottomrule
\end{tabular}
}
\caption{Overview of the language models used in the experiments, including their developers and access types.}
\label{tab:llm_overview_general}
\end{table}

\begin{table}[h]
\centering
\resizebox{\columnwidth}{!}{
\begin{tabular}{llll}
\toprule
\textbf{Model} & \textbf{Parameters} & \textbf{Architecture} & \textbf{Context} \\ 
\midrule

GPT-5.1
& Undisclosed & Proprietary & 400K \\

GPT-5-mini
& Undisclosed & Proprietary & 400K \\

Gemma-4-31B-IT~\cite{gemmateam2026gemma4}
& 31B & Dense & 256K \\

Gemma-4-26B-A4B-IT~\cite{gemmateam2026gemma4}
& 25.2B (3.8B active) & MoE & 256K \\

Gemma-4-12B-IT~\cite{gemmateam2026gemma4}
& 12B & Dense & 256K \\

Gemma-4-E4B-IT~\cite{gemmateam2026gemma4}
& $\sim$4.5B effective & Dense & 128K \\

Llama-3.1-70B~\cite{grattafiori2024llama3herdmodels}
& 70B & Dense & 128K \\

Llama-3.1-8B~\cite{grattafiori2024llama3herdmodels}
& 8B & Dense & 128K \\

Qwen3-32B~\cite{qwen3technicalreport}
& 32B & Dense & 32K$^{*}$ \\

\bottomrule
\end{tabular}
}
\caption{Model specifications, including parameter counts, architectures, and context-window sizes.}
\label{tab:llm_overview_specs}
\end{table}

\section{Evaluation Prompts}
\label{appendix:prompts}

Table \ref{tab:system_prompts} lists the system prompts used for evaluation. 

\begin{table*}[h]
    \centering
    \renewcommand{\arraystretch}{1.5} 
    \begin{tabular}{|p{0.25\textwidth}|p{0.68\textwidth}|}
        \hline
        \textbf{Prompt Title/Name} & \textbf{Prompt} \\
        \hline
        
        \textbf{System Prompt A: Inheritance Defeasible Reasoning} & 
        You are an expert in defeasible reasoning. You will be given a category hierarchy where entities belong to categories, and categories can belong to other categories. A property that applies to a category generally applies to all entities and subcategories within it, unless explicitly stated otherwise for a specific entity or subcategory. \newline\newline
        All previously given information remains active throughout the conversation. You will then receive a series of information updates one at a time. After each update you will be asked whether all the information so far supports the hypothesis. \newline\newline
        Answer \textbf{yes} if the hierarchy and facts support the hypothesis and nothing blocks inheritance for the specific entity in question, \textbf{no} if inheritance is explicitly blocked for that entity, or \textbf{unknown} if you cannot determine either way. \newline\newline
        Output only the answer word and nothing else. Do not explain, justify, or add any additional text. \\
        \hline
        
        \textbf{System Prompt B: Defeasible Reasoning with Effect Turns} & 
        You are an expert in defeasible reasoning. You will be given a set of general rules and specific facts about objects. The rules are defaults: they hold generally unless specific information contradicts them for a particular object. All previously given information remains active unless explicitly overridden. \newline\newline
        You will then receive a series of information updates one at a time. After each update you will be asked one of two questions:

        \textbf{Support:} Does the information support the hypothesis? Answer \textbf{yes}, \textbf{no}, or \textbf{unknown}.
        
        \textbf{Effect:} Does the update \textbf{strengthen}, \textbf{weaken}, or have \textbf{no effect} on the support for the hypothesis?

        When information comes from multiple sources, rely on the most reliable source when sources contradict each other. If a less reliable source makes a claim and a more reliable source makes a different but non-contradicting claim, both claims remain active independently. A more reliable source speaking does not undo what a less reliable source established unless they are making opposite claims about exactly the same fact. \newline\newline
        Output only the answer word and nothing else. Do not explain, justify, or add any additional text. \\
        \hline
    \end{tabular}
    \caption{System Prompts for Evaluation}
    \label{tab:system_prompts}
\end{table*}

\section{Resource Information} 
\label{appendix:resources}

Table \ref{tab:llm_overview_general} and \ref{tab:llm_overview_specs} cover the information about LLMs used in experiments. We evaluate nine base models under ten model–inference configurations, including standard and extended-thinking variants of Qwen3-32B.
For proprietary/closed models (GPT-5.1 and GPT-5-mini), we kept the temperature at 0.1. For open-weighted models, we used constrained decoding. We ran 3 A40 GPUs ($48\times3=144$ GB) for approximately 180 hours. 

We used Claude Code \footnote{https://claude.com/product/claude-code} for coding purposes and ChatGPT \footnote{https://chatgpt.com/} for refining our writing.

\end{document}